\documentclass{article}
\usepackage{ijcai23}

\usepackage{times}
\usepackage{soul}
\usepackage{url}
\usepackage[hidelinks]{hyperref}
\usepackage[utf8]{inputenc}
\usepackage[small]{caption}
\usepackage{amsmath,amssymb,amsfonts,amsthm}
\usepackage{float}
\usepackage{appendix}
\usepackage{graphicx}
\usepackage{subfigure}
\usepackage{booktabs}
\usepackage{color}
\usepackage[algoruled,boxed,lined]{algorithm2e}
\DeclareMathOperator*{\argmax}{argmax}
\usepackage{algorithmic}
\usepackage[switch]{lineno}

\title{BARA: Efficient Incentive Mechanism with Online Reward Budget Allocation in Cross-Silo Federated Learning}

\author{
    Yunchao Yang$^{1,2}$, \, Yipeng Zhou$^{3}$, \, Miao Hu$^{1,2}$ \footnote{Corresponding author.}, \, Di Wu$^{1,2}$, \, Quan Z. Sheng$^{3}$
    \affiliations
    $^{1}$School of Computer Science and Engineering, Sun Yat-sen University, Guangzhou, China
    \affiliations
    $^{2}$Guangdong Key Laboratory of Big Data Analysis and Processing, Guangzhou, China
    \affiliations
    $^{3}$School of Computing, Faculty of Science and Engineering, Macquarie University, Sydney, Australia
    \emails
    yangych65@mail2.sysu.edu.cn
    \emails
    \{ humiao5, wudi27\}@mail.sysu.edu.cn, \{yipeng.zhou, michael.sheng\}@mq.edu.au
}

\begin{document}

\maketitle

\begin{abstract}

Federated learning (FL) is a prospective distributed machine learning  framework that can preserve data privacy. In particular, cross-silo FL can complete model training by making isolated data islands of different organizations collaborate with a parameter server (PS) via exchanging model parameters for multiple communication rounds. In cross-silo FL, an incentive mechanism is indispensable  for motivating data owners to contribute their models to FL training. However, how to allocate the reward budget among different rounds is an essential but complicated problem  largely overlooked by existing works. The challenge of this problem lies in the opaque feedback between reward budget allocation and model utility improvement of FL, making the optimal reward budget allocation complicated. To address this problem, we design an online reward budget allocation algorithm using Bayesian optimization named BARA (\underline{B}udget \underline{A}llocation for \underline{R}everse \underline{A}uction). Specifically, BARA can model the complicated relationship between reward budget allocation and final model accuracy in FL based on historical training records so that the reward budget allocated to each communication round is dynamically optimized so as to maximize the final model utility. We further incorporate the BARA algorithm into reverse auction-based incentive mechanisms to illustrate its effectiveness. Extensive experiments are conducted on real datasets to demonstrate that BARA significantly outperforms competitive baselines by improving  model utility with the same amount of reward budget.

%

%the client's local data from being directly exposed to the parameter server (PS). However, most of existing works generally assume that the client is willing to participate in FL, which is unrealistic in practice. Incentive mechanism aims to motivate more high-quality clients to engage in FL. Reverse auction is one of the most widely used basic theories in the incentive mechanism design of FL. Nevertheless, the existing reverse auction-based incentive mechanisms do not consider how much reward budget should be allocated to each auction under the limited total reward budget, which will cause the final model to fail to achieve satisfactory performance. In this paper, we propose an  using Bayesian optimization for reverse auction-based incentive mechanism of FL. We formalize the reward budget allocation problem as a Multi-armed Bandit (MAB) problem. An approximating final model accuracy method is designed to obtain observation, which is employed to guide the trade-off between exploration and exploitation in Bayesian optimization. To our best 
 %knowledge, we are the first to study the reward budget allocation problem in reverse auction-based incentive mechanisms of FL. Experimental results demonstrate the effectiveness of BARA, and it is orthogonal to existing reverse auction-based incentive mechanisms of FL, which means BARA can be directly used to improve the performance of existing methods.

\end{abstract}

\section{Introduction}

Due to the rising concern on data privacy leakage in recent years, laws such as General Data Protection Regulation (GDPR) \cite{voigt2017eu} have been made to regulate the collection and use of user data to protect data privacy. Federated learning (FL) \cite{McMahan2016CommunicationEfficientLO}, as an emerging distributed  machine learning paradigm, enables decentralized clients to collaboratively train a shared model  without disclosing their private data. The workflow of FL mainly includes: 1) The parameter server (PS) distributes the latest global model to participating clients. 2) Each client updates the model with its local dataset  and returns the updated model to the PS. 3) Model aggregation is performed on the PS to obtain a new global model for the next communication round. The above process is repeated until the maximum number of training rounds is reached. Due to its capability in preserving data privacy, the FL market is proliferating. According to \cite{flbusinessvalue}, the global FL market size is projected to increase from 127 million dollars in 2023 to 210 million dollars by 2028, at a compound annual growth rate of 10.6\% during the forecast period.

However, FL is unsealed without a mechanism to force clients to altruistically contribute their models, and thus an effective incentive mechanism motivating clients is very essential for the success of FL. It was reported in \cite{ng2021multi} that the final model accuracy of FL can be substantially improved by an incentive mechanism, which can inspire more high-quality clients to participate in FL. Reverse auction \cite{myerson1981optimal}  has been widely studied in the incentive mechanism design for FL. For example, Deng \emph{et al.} \cite{deng2021fair} designed a quality-aware incentive mechanism based on reverse auction to encourage the participation of high-quality learning users. RRAFL \cite{zhang2021incentive} is an incentive mechanism for FL based on reputation and reverse auction theory. A few works even attempted to design incentive mechanisms for FL enhanced by differential privacy. For example, FL-Market \cite{zheng2021incentive} was proposed as  a novel truthful auction mechanism enabling data owners to obtain rewards according to their privacy losses quantified by local differential privacy (LDP).

Yet, existing works focused on how to optimally allocate rewards between heterogeneous participating clients, ignoring the essential problem of how to allocate rewards between communication rounds. It usually consumes a large number of communication rounds  to train advanced machine learning models. 
On the one hand, if excessive rewards are allocated per communication round, the reward budget will be used up instantly without fully utilizing clients' data for model training.  On the other hand, if the amount of allocated rewards is insufficient to solicit high-quality clients, FL fails as well because of the low training efficiency per round. 
%, no work has yet considered the reward budget allocation problem in the reverse auction-based incentive process. Generally speaking, there will be an auction in each communication round and the reward budget of the PS in each auction determines how many clients will be recruited in FL. There is a total reward budget in PS, which is allocated evenly to each auction in existing works. However, allocating a larger reward budget to each auction will lead to a reduction in the total communication rounds. In our work, we focus on how to make a trade-off between the number of participating clients in each communication round and the total communication rounds.

How to optimally allocate the reward budget between communication rounds is extremely difficult because of the complicated relationship between  the final model accuracy and  reward budget allocation giving rise to the following three challenges. \emph{First}, once we change the amount of reward budget per communication round, it yields two opposite influences. If we increase the amount of reward budget per communication round, it diminishes the total number of conducted rounds but increases the model utility improvement per communication round, and vice verse. Thereby, it becomes vague how the change of the reward budget allocation will eventually affect the change of final model utility. \emph{Second}, the prediction is problem-related susceptible to random factors, which will be different when training different models using different datasets or hyperparameters (\emph{e.g.}, learning rate and batch size). \emph{Third}, due to the constrained total reward budget, it is unaffordable to exhaustively make trials with different reward budget allocation strategies to search for the best one.

%\begin{itemize}
 %   \item \textbf{The objective function is unknown and expensive to evaluate.} Our goal is to maximize the final model accuracy by balancing the number of participating clients in each communication round and the total communication rounds. We have no way to accurately quantify the impact of these two variables on the final model accuracy. Moreover, it is expensive to evaluate the value of objective function since completing one round of FL training process consumes a lot of communication and compute resources \cite{kairouz2021advances}.
  %  \item \textbf{The observations are noisy and cannot be obtained directly.} Given a fixed total reward budget, we aim to find an optimal reward budget allocation setting that maximizes the final model accuracy. However, it is impractical to obtain the final model accuracy before the entire FL workflow is over, which means we can only observe the noisy model accuracy of current communication round after the PS decides the reward budget for current auction. And, the inherent non-IID problem \cite{zhao2018federated} of FL aggravates the gap between real value and observed value.
%\end{itemize}

To address the reward budget allocation problem, we  design a novel online reward budget allocation algorithm named BARA (\underline{B}udget \underline{A}llocation for \underline{R}everse \underline{A}uction). Specifically, a Gaussian process (GP) model \cite{williams2006gaussian} is established to analyze relationship between reward budget allocation  and final model utility.  Newton's polynomial interpolation is  used to expand the  training records for establishing the GP model based on historical information.  
Bayesian optimization is employed to  dynamically search the reward 
 budget allocation strategy that can  maximize the predicted final model utility.  Based on the above analysis, we design the BARA algorithm to  determine reward budget allocation per round
in an online fashion.  It is worth noting that BARA is orthogonal to existing incentive mechanisms optimizing reward allocation between clients. Thus, it can be widely incorporated into existing incentive mechanisms to boost FL utility, which is exemplified by incorporating BARA into reverse auction-based incentive mechanisms in our work.  %We conduct extensive experiments on four public datasets to evaluate our algorithm in comparison with competitive baselines. The results demonstrate the effectiveness of BARA in improving model utility. %In addition, BARA is orthogonal to existing reverse auction-based incentive mechanisms of FL and can be used to improve the performance of existing methods

%for of FL. We first formulate the reward budget allocation problem as a Multi-armed Bandit (MAB) problem and transforms the problem into deciding the number of participating clients in each communication round, which establishes a connection between the reward budget and the number of participating clients. Then we design an approximating final model accuracy method based on Newton's polynomial interpolation, which is regarded as the observation after the PS make a decision. After that, we relationship between exploration and exploitation. The Bayesian posterior will be updated based on new observations to assist the PS to learn the optimal decision. We also separates BARA into two stage: a pure exploration stage in which the PS randomly selects the number of participating clients, followed by an exploration-exploitation stage in which Bayesian optimization is performed with the enriched prior knowledge. Moreover, we differentiate stale and fresh observations during Bayesian posterior update to mitigate noise disturbance.

In summary, our main contributions are presented as follows:

\begin{itemize}
   \item   We establish a GP model to analyze the relation between reward budget allocation and model accuracy. Newton's polynomial interpolation is applied to enrich the training records for the GP model while the Bayesian optimization is employed to search the optimal reward budget allocation strategy.
   \item Based on our analysis, we propose an online reward budget allocation algorithm (BARA). To our best knowledge, we are the first to address the reward budget allocation across multiple communication rounds to incentivize FL clients.
  % Based on historical records, we enrich training records using . Then,  with enriched knowledge.
   \item We conduct extensive experiments on four public datasets to evaluate our algorithm in comparison with other baselines. The results demonstrate the extraordinary performance and application value of BARA. %Moreover, BARA is orthogonal to existing incentive mechanisms for FL and can be used to improve the performance of existing methods.
\end{itemize}

\section{Related Work}
\label{sec:related_work}

In this section, we briefly discuss related works on incentive mechanism design for FL and  Bayesian optimization.

\subsection{Incentive Mechanisms in Federated Learning}

Incentive mechanism design encouraging clients to contribute their resources for conducting FL has attracted intensive research  in recent years. FAIR \cite{deng2021fair} was proposed as an incentive mechanism framework in which   reverse auction  was employed to incentivize clients by rewarding them based on their model quality and bids.  Clients contributing  non-ideal model updates will be filtered out before the model aggregation stage. Zeng \emph{et al.} \cite{zeng2020fmore} proposed an incentive mechanism  with multi-dimensional procurement auction of winner clients.  Theoretical results of this strategy was  provided as well. 
RRAFL \cite{zhang2021incentive} was designed by combining reputation and reverse auction theory together to reward participants in FL. A reputation calculation method was proposed to measure the clients' quality. Zhou \emph{et al.} \cite{zhou2021truthful} considered how to guarantee the completion of FL jobs with minimized  social cost. They decomposed the problem into a series of winner determination problems, which were further solved by reverse auction.

It is a more challenging problem to incentivize clients in differentially private federated learning (DPFL) due to the disturbance of noises. FL-Market \cite{zheng2021incentive} is a personalized  LDP-based FL framework with auction to incentivize the trade of private models with less significant noises. It can  meet diversified privacy preferences of different data owners % (i.e., risk-tolerant data owners and \emph{risk-averse} data owners) 
before deciding how to allocate rewards. Liu \emph{et al.} \cite{liu2021privacy} introduced the cloud-edge architecture into FL incentive mechanism design to enhance privacy protection.
Sun \emph{et al.} \cite{sun2021pain} proposed a contract-based personalized privacy-preserving incentive mechanism for FL by
customizing a client's reward as the compensation for privacy leakage cost.% with the help of privacy protection level.

% According to state-of-the-art  works, reverse auction is the most fundamental technique underlying  FL incentive mechanism design.

Nevertheless, existing works failed to optimize the reward budget allocation across multiple communication rounds due to the difficulty to explicitly analyze the relationship between a reward budget allocation strategy  and final model utility, and this problem will be initially investigated by our work.

\subsection{Bayesian Optimization}

Bayesian optimization based on Gaussian process (GP) is particularly effective in analyzing a complicated process susceptible to various random factors without  the need to derive  a closed-form solution. 
%ng objective functions  powerful strategy for finding the extrema of objective functions that are expensive to evaluate. It is applicable in situations where one does not have  for the objective function, but where one can obtain observations (possibly noisy) of this function at sampled values .
Based on
kernel methods and GP models, significant contribution has been made in machine learning  \cite{shahriari2015taking}.
In \cite{williams2006gaussian}, smoothness
assumptions of the objective function to be modelled are encoded through flexible kernels  in a  nonparametric fashion. 
Srinivas \emph{et al.} \cite{Srinivas2009GaussianPO} proposed GP-UCB, an intuitive upper-confidence based algorithm. Its cumulative regret in terms of maximal information gain was bounded, and a novel connection between GP optimization and experimental design was established. 
Bogunovic \emph{et al.} \cite{bogunovic2016time} considered a sequential Bayesian optimization problem with bandit feedback, which set the reward function to vary with time. 
GP-UCB was extended to provide an explicit characterization of the trade-off between the time horizon and the rate at which the function varies.
% Practical applications may be subject to a variety of safety constraints implying that actions cannot be freely chosen from the entire action space, which has been considered by following works. 
% SAFEOPT \cite{sui2015safe} was proposed as a safe optimization algorithm. Its convergence to a natural notion of optimum  under safety constraints was theoretically guaranteed. STAGEOPT \cite{sui2018stagewise} separated the
% safe optimization problem into safe expansion and utility
% optimization phases. 
% However, STAGEOPT is much more amenable to a variety of related settings than SAFEOPT.

Given the excellent performance of Bayesian optimization based on GP  in modeling complicated processes influenced  by multiple unknown random factors, our work is novel in applying this approach in FL incentive mechanism design. %  to model the complicated relationship between user 

%In reality, performing FL training  consumes both  computing and communication resources. Given a limited total reward budget to compensate consumed resources, how to optimally allocate reward budget across different training rounds is still an open problem,  %To our best knowledge, we are the first to employ Bayesian optimization to perform reward budget allocation between different communication rounds. 

%solve the online reward budget allocation problem. 

%model but in the setting of reverse auction-based incentive mechanism in FL. 

%And previous works did not take into account the prediction difficulty of observations and the non-IID nature of FL, which will be addressed in our design.

\section{Preliminaries}
\label{sec:preliminaries}

We investigate a generic FL system with  a single parameter server (PS) owning the test dataset $\mathcal{D}_{test}$ and $N$ clients with local datasets $\left\{\mathcal{D}_{1}, \mathcal{D}_{2}, \dots, \mathcal{D}_{N} \right\}$. 
FL training is completed by multiple communication rounds denoted by rounds $\left\{1, 2, \dots, t, \dots, T \right\}$.
In communication round $t$, a typical FL system with a reverse auction-based incentive mechanism \cite{deng2021fair} works as follows: %In each communication round $t$, the workflow is as follows: 
\begin{itemize}
    \item {\bf Step 1 (on clients):} 
    Client $i$ reports its bid $b_{t,i}$ to the PS,  representing the reward client $i$ desires for participating in FL.
    \item {\bf Step 2 (on the PS):}
    The PS measures the quality of  each client (\emph{e.g.}, the size of local dataset)   denoted by $q_{t,i}$. Based on $q_{t,i}$  and  $b_{t,i}$, the PS selects $n_t$ ($1 \le n_t <N$) as participating clients %$n$ ($1 \le n<N$) and whether client $i$ will be selected, 
    and $r_{t,i}$ represents the reward allocated to client $i$. Then, the PS sends out the latest global model $\omega_{t-1}$ to participating clients.
    \item {\bf Step 3 (on participating clients):} 
    Each participating client updates $\omega_{t-1}$ with its local dataset by using local update algorithm (\emph{e.g.}, the gradient descent algorithm \cite{ruder2016overview}). Then, model updates are returned to the PS.
    \item {\bf Step 4 (on the PS):} 
    The PS performs model aggregation based on returned model updates to obtain the global model $\omega_{t}$ for the next communication round.
\end{itemize}

To effectively incentivize clients, reverse auction is widely adopted to determine the reward allocated between clients. 
%Then we elaborate the strategy of client selection and reward allocation in reverse auction. 
More specifically, the PS ranks all clients in terms of the ratio of quality over bid, i.e.,  $\frac{q_{t,i}}{b_{t,i}}$, in a descending order. Then, the PS can get a list of ranked clients with $\frac{q_{t,1}}{b_{t,1}}>\frac{q_{t,2}}{b_{t,2}}> \dots >\frac{q_{t,N}}{b_{t,N}}$. 
Due to the limited total reward budget, the PS sets a reward budget limit $B_t$ to reward  participating clients in communication round $t$. Constrained by $B_t$, the PS selects top clients from the rank list until $B_t$ is used up. Suppose top $n_t$ clients are selected, the reward for client  $i$ ($1 \le i \le n_t$) is $r_{t,i}=\frac{b_{t,n_t+1}}{q_{t,n_t+1}}q_{t,i}$. 
%The value of $n$ can be easily determined by the  constraint of limited total reward budget $B_{t}$ which gives:
$n_t$ is determined by the constraint of $B_t$:
\begin{align}
\label{eq:limit_participants_number}
    n_t = \argmax_{n} \sum_{i=1}^{n}\frac{b_{t,n+1}}{q_{t,n+1}}q_{t,i} \leq B_{t}.
\end{align}

A straightforward strategy to determine $B_t$ adopted by existing works is to set the target number of communication rounds $T_{max}$. If the total reward budget is $B_{total}$, the reward budget for each communication round is 
$B_{t}=\frac{B_{total}}{T_{max}}$. % denote the total reward budget of the PS and the maximum number of communication rounds, respectively. Existing works simply allocate $B_{total}$ evenly into each communication round (i.e., $, which does not consider the impact of the number of participating clients and the total communication rounds on the final global model accuracy and this problem will be addressed in our work.

\section{Problem Formulation}
\label{sec:problem_formulation}

In this work, we design a novel algorithm to adjust the number of participating clients  based on the typical workflow of FL system with a reverse auction-based incentive mechanism in Section \ref{sec:preliminaries}. 

Let $B_{total}$  denote the total reward budget provided by the PS to recruit participating clients. Let $a_t$ denote the model accuracy after communication round $t$
 and $\Delta a_t$ denote the incremental improvement of model accuracy, i.e., 
 $\Delta a_t = a_t-a_{t-1}$. $n$ ($1 \le n < N$) is the number of participating clients. Our objective is to tune the number of recruited clients to adjust consumed reward budget per communication round so as to maximize the final model accuracy. If $T$ communication rounds are conducted in total, our problem can be formulated as:
\begin{align}
   \mathbb{P}1 \quad & \max_{n:1\leq  n<N}  a_{0}+\sum_{t=1}^{T}\Delta a_{t},\\
    s.t. ~ 
    & \sum_{t=1}^{T}B_{t} \le B_{total},
\end{align}
 %In each communication round, the PS has to spend a higher reward budget to have more clients participate in FL. However, the more reward budget $B_{t}$ the PS consumes in each communication round, the less the total communication rounds will be since tis limited. The reward budget allocation problem can be transformed into deciding the number of participating clients in each communication round. Our goal is to strike a balance between the number of participating clients per communication round and the total communication rounds to maximize the final global model accuracy. In this paper, we will address the optimization problem formulated blow:
where $a_{0}$ is the test accuracy of the initial global model $\omega_{0}$. However, it is a very challenging problem because: 1) $\Delta a_t$ is a function of $n$ as a bigger $n$ brings a larger $\Delta a_t$. 2) $B_t$ is a function of $n$ since more participating clients consume more reward budget. 3) $T$ is a function of $n$ as well given a fixed $B_{total}$. Thus, if these variables in $\mathbb{P}1$ are expressed as functions of $n$, we can get: 
\begin{align}
   \mathbb{P}2 \quad & \max_{n:1\leq  n<N}  a_{0}+\sum_{t=1}^{T(n)}\Delta a_{t, n},\\
    s.t. ~ 
    & \sum_{t=1}^{T(n)}B_{t}(n) \le B_{total}.
\end{align}

With the knowledge of $T(n)$, $\Delta a_{t, n}$ and $B_t(n)$, we can solve $\mathbb{P}2$. %We assume that each client's bid is independently sampled from a uniform distribution in each communication round, which is the same as \cite{deng2021fair,zhang2021incentive}.
The term $B_{t}(n)$ is determined by the reverse auction-based incentive mechanism once $n$ is fixed. As discussed in Section \ref{sec:preliminaries}, we can calculate the minimum reward budget $B_{t}(n)$ consumed by selecting $n$ clients to participate in FL in the $t$-th communication round as:
\begin{align}
    B_{t}(n)=\sum_{i=1}^{n}\frac{b_{t,n+1}}{q_{t,n+1}}q_{t,i}.
\end{align}

Based on  $B_{t}(n)$, $T(n)$ can be computed correspondingly. %The number of participating clients in the $\tau$-th communication round ($1 \le \tau \le t$) is denoted by $n_{\tau}$. 
Let $x_{\tau,n}=1$ denote if there are $n$ participating clients in communication round $\tau$. %We have $\sum_{n=1}^{N-1}x_{\tau,n}=1$ for each communication round $\tau$. Therefore, 
$T(n)$ can be estimated if $n$ participating clients are selected in communication  round  $t$ as:
\begin{align}
    T_t(n) \approx \frac{B_{total}}{\bar{B}_{t}(n)},
\end{align}
where $\bar{B}_{t}(n) =\frac{\sum_{\tau=1}^{t}x_{\tau,n}B_{t}(n)}{\sum_{\tau=1}^{t}x_{\tau,n}}$ is the average reward budget consumption per communication round estimated in  round $t$. 
Unfortunately, there is no prior work that explicitly defines $\Delta a_{t, n}$. Thus, the main challenge for solving  $a_{t, n}$ is how to accurately  estimate $\Delta a_{t,n}$.

\section{Methodology}
\label{sec:methodology}

In this section, we propose an online reward budget allocation algorithm using Bayesian optimization to solve  $\mathbb{P}2$. We first utilize Newton's polynomial interpolation to synthesize training records based on historical records observing reward budget allocation 
and model accuracy improvement. Based on training records, a Gaussian process (GP) is established to model the relationship between final model accuracy and  reward budget allocation strategies. 
 Next,  Bayesian optimization is employed to search for the optimal reward budget allocation strategy in an online fashion to maximize the final model accuracy.

%incentive mechanism and the pipeline of system is shown in Fig. \ref{fig:bara_global_view}.

% \subsection{Enriching Training Records}
\subsection{Training Records Synthesis} 

How to exactly estimate $\Delta a_{t, n}$ (representing model accuracy improvement with $n$ participating clients in the $t$-th communication round) is a challenging open problem. When training different models, we can get different $\Delta a_{t, n}$.  Until communication round $t$, we denote the number of participating clients in the $\tau$-th communication round as $n_{\tau}$ ($1 \le \tau \le t$). However, we cannot compute $\Delta a_{\tau, n}$ prior to model training and it is also impossible to obtain $\Delta a_{\tau, n}$ if $n \neq n_{\tau}$. Thereby, we approximate unknown training records during the model training process with historical records of $\Delta a_{\tau, n}$. We can use a matrix $\mathbf{M}_{t}\in \mathbb{R}^{t\times N-1}$ to denote $\Delta a_{\tau, n}$ until communication round $t$ (i.e., $\mathbf{M}_{t}=[\Delta na_{\tau,n}]$ ($1 \le \tau \le t$ and $1 \le n <N$). $\Delta a_{\tau,n}$ will be empty if $n \neq n_{\tau}$. Each row represents a communication round and note that only a single element in each row is from  FL training records since we can only select a single $n_{\tau}$ for communication round $\tau$.

However, to determine which $n$ yields the highest final model accuracy, we need the knowledge of all elements in $\Delta a_{1,n}, \Delta a_{2,n}, \dots, \Delta a_{T(n),n}$. With all known elements in the $n$-th column of  $\mathbf{M}_{t}$, we employ the Newton's polynomial interpolation \cite{hildebrand1987introduction} to approximate the values of those unknown elements. % with the same $n$. 

%accuracy in the $T(n_{t})$-th communication round. Take the $n_{t}$-th column from the matrix $M_{t}$.

% {\bf YP: reference is needed for citing this method. }
Without loss of generality, we briefly explain how to  apply the Newton's polynomial interpolation  for a particular $n$-th column. 
%We assume $\sum_{\tau=1}^{t}x_{\tau,n_{t}}=J$, 
Let $t_1, t_2, \dots, t_J$ denote indices of elements in the $n$-th column with known value from past training records. 
%$\left\{t_{j} \right\}_{j=1}^{J}$ and $\left\{\Delta a_{t_{j},n_{t}} \right\}_{j=1}^{J}$ denote all communication rounds $t_{j}$ where $x_{t_{j},n_{t}}=1$ and the corresponding accuracy improvement $\Delta a_{t_{j},n_{t}}$, respectively. 
Then, we can define a number of divided differences as follows:
\begin{align}
\label{EQ:NewTon}
    y[t_{1},t_{2},\cdots ,t_{J}]=\frac{y[t_{1},\cdots ,t_{J-1}]-y[t_{2},\cdots ,t_{J}]}{t_{1}-t_{J}}.
\end{align}

Here $y[t_{j}]=\Delta a_{t_{j},n}$ for all $1 \le j \le J$. We can easily compute $ y[t_{1},t_{2},\cdots ,t_{J}]$ with $y[t_{j}]$ and Eq.~\eqref{EQ:NewTon}. Unknown values in  $\mathbf{M}_t$ until $t=T(n) $ can be estimated by:
\begin{align}
\label{EQ:EstDA}
    \Delta \hat{a}_{\tau,n}=
    & y[t_{1}]+y[t_{1},t_{2}](\tau -t_{1})+\cdots \nonumber \\
    & +y[t_{1},t_{2},\cdots ,t_{J}]\prod_{j=1}^{J-1}(\tau -t_{j}),
\end{align}
for $\tau =1, 2, \dots, t, \dots, T(n)$. Let $T = \max_{1 \le n<N} T(n)$ denote the maximum number of communication rounds we can conduct by selecting $n$ clients per communication round. Based on interpolation results, we can create the estimation matrix $\widehat{\mathbf{M}}_{T}=[\Delta \hat{a}_{t,n}]$ ($1 \le t \le T$ and $1 \le n <N$). Note that $\Delta \hat{a}_{t,n}$ is valid only if  $t \leq T(n)$. Although $\Delta \hat{a}_{1, n}, \dots, \Delta \hat{a}_{T(n), n}$ can be estimated through Eq.~\eqref{EQ:EstDA}, the predicted model accuracy improvement is vulnerable to overfitting. In particular, in the first few communication rounds, the number of available historical records is insufficient for accurately finding the best $n$.  Thus, we establish  a learning process to dynamically predict the accuracy with different $n$ before we can decide the optimal number of participating clients.

In the next subsection, we use a Gaussian process (GP) to model the change of final model accuracy with different reward budget allocation strategies. Based on GP, the Bayesian optimization technique is further applied to determine the optimal $n$ for each communication round.

%To avoid overfitting and efficiently explore the performance improvement with different $n$'s, 
% In the next subsection, we use a Gaussian process (GP) to model the change of final model accuracy with different reward budget allocation strategies. Based on GP, the Bayesian optimization technique is further applied to determine the optimal $n$ for each communication round. 

%After that, the final global model accuracy $a_{T(n_{t}),n_{t}}$ (abbreviated as $a_{n_{t}}$) can be approximated as:
%\begin{align}
 %   \label{con:cal_final_acc}
%  a_{n_{t}}=a_{0}+\sum_{\tau=1}^{T(n_{t})} \Delta a_{\tau,n_{t}}.
%\end{align}

%Here we can get the historical records of the number of participating clients $\mathbf{n}_{t}=\left\{n_{\tau} \right\}_{\tau=1}^{t}$ and the approximated final global model accuracy $\mathbf{a}^{t}=\left\{a_{n_{\tau}} \right\}_{\tau=1}^{t}$, which will be used to guide the PS's decision in the next communication round.

\subsection{Searching for Optimal Reward Budget Allocation Strategy}

It is known that the model accuracy performance is susceptible  to various factors such as the data distribution among clients. It is difficult to accurately predict final model accuracy only based on the number of participating clients. In light of this complication, we use a Gaussian process (GP) to model the random evolution of final model accuracy when taking different reward budget allocation strategies.

%z_t(n): represent final model accuracy from t=1 until end  n predicted at round t. 

%\mathbbf{z}_t is a vector of z_t(n_t), which obeys eq 12. 

%\epsilon_t follows the gap between f(n) and z_t(n), which obeys eq 13, 14 

As we collect more records of $n_t$ and $\Delta \hat{a}_{t, n_t}$ along the training process, the matrix   $\widehat{\mathbf{M}}_t$ expands gradually and the predicted value of each unknown element will be updated according to Eq.~\eqref{EQ:EstDA}. Note that our goal is to predict the final model accuracy when using different $n$. For convenience, we define $\hat{a}_t(n)$ to represent the estimated final model accuracy predicted at communication round $t$. In other words,  $\hat{a}_t(n) = a_0 +\sum_{t=1}^{T(n)} \Delta \hat{a}_{t, n}$ where $\Delta \hat{a}_{t, n}$ are elements in the $n$-th column of $\widehat{\mathbf{M}}_{T}$.

Note that $\hat{a}_t(n)$ is derived based on a few observations in matrix $\widehat{\mathbf{M}}_{T}$. It only utilizes elements in the $n$-th column for prediction failing to fully utilize all observations to predict   $\hat{a}_t(n)$. To overcome this drawback, we model  $\hat{a}_t(n)$ with a GP to capture the relationship between  $\hat{a}_t(n)$ and  $\hat{a}_{t'}(n')$ so that we can fully utilize all observations to more accurately predict $\hat{a}_t(n)$.  

%In the $t$-th communication round, the PS selects $n_{t}$ participating clients and obtain the observation $a_{n_{t}}$, which has been approximated properly. 
%The deviation between the observed value and the real value can be measured by a zero mean random noise: $\epsilon_{t}\sim \mathcal{N}(0,\sigma^{2})$, which is independent at different time. In general, it is extremely expensive to evaluation the objective function since each round of FL training process consumes a lot of communication and compute resources \cite{kairouz2021advances}. Fortunately, Bayesian optimization techniques are some of the most efficient approaches in terms of the number of function evaluations required. The prior will be updated in light of new observations to get the posterior.  Here we define the prior distribution over the objective function as a Gaussian Process (GP) \cite{williams2006gaussian}, which is the common choice for the prior distribution of Bayesian optimization \cite{mockus1994application}. 

%until  round t
%f1(n1), f2(n2), ... ft(nt)
%mu_1 n1, mu2_n2......
%kn1,n1,  kn2,n2,.....
%f_t(n) -> mun, 
%var  -> knn

%'  N variances

To distinguish with final model accuracy obtained by interpolation via Eq.~\eqref{EQ:EstDA}, we define $z_t(n)$ as the final model accuracy sampled from the GP  model in the $t$-th communication round assuming that $n$ clients are recruited to conduct FL in each communication round. Specifically,  $z_t(n_t)$ is modeled as a random variable sampling values  from the distribution of GP($\mu(n_t), k(n_t,n_t)$) where $k(n_t,n_t)$ is the covariance (or kernel) function. The  mean value of $z_t(n_t)$ is denoted by $\mu(n_t)=\mathbb{E}[z_t(n_t)]$. With multiple variables $z_t(n_t)$, we needs to consider the covariance $\mathbb{E}[(z_t(n_t)-\mu(n_t))(z_t(n_{t'})-\mu(n_{t'}))]$ when modeling the relationship between two choices of $n_t$ and $n_{t'}$ at two different communication rounds $t$ and $t'$, respectively.

% and covariance (or kernel) function $k(n,n')=\mathbb{E}[(f(n)-\mu(n))(f(n')-\mu(n'))]$, where smoothness assumptions about f are encoded through the choice of kernel in a flexible nonparametric fashion. We use the squared exponential kernel as our covariance function, which is a common choice for the covariance function%$\mu(n)=\mathbb{E}[f(n)]$ and covariance (or kernel) function $k(n,n')=\mathbb{E}[(f(n)-\mu(n))(f(n')-\mu(n'))]$, where smoothness assumptions about f are encoded through 
%The choice of kernel in a flexible nonparametric fashion. 

According to \cite{Srinivas2009GaussianPO}, the squared exponential kernel function is widely adopted to model covariance for a GP. Note that the elements in $\widehat{\mathbf{M}}_{t}$ (i.e., $\Delta \hat{a}_{t,n}$) are approximated by Newton's polynomial interpolation. Therefore, the approximation gets better over time as we collect more observation records of $\Delta a_{t, n}$, which means fresh observations are more valuable than stale ones. We construct the composite kernel to weigh stale and fresh observations differently based on Ornstein-Uhlenbeck temporal covariance function. Together with the squared exponential kernel function, the covariance between $z_t(n_t)$ and $z_{t'}(n_{t'})$ is  modeled by
\begin{align}
    k(n_t,n_{t'})=(1-\lambda)^{\frac{|t-t'|}{2}} \exp \left ( -\frac{||n_t-n_{t'}||^{2}}{2l^{2}} \right ),
\end{align}
where $l > 0$ is a length scale hyperparameter to determine how much the two points $n_t$ and $n_{t'}$ influence each other. Intuitively, if $n_t$ is closer to $n_{t'}$, the value of $k(n_t,n_{t'})$ is bigger implying that $z_t(n_t)$ and $z_{t'}(n_{t'})$ are more correlated. Thus, the information  of  $z_t(n_t)$ is more useful for us to predict $z_{t'}(n_{t'})$. Moreover,  stale observations should be weighted lighter and lighter over time. Here $\lambda$ controls how fast the weights of stale observations decrease.

\begin{algorithm}[t]
\SetNoFillComment
\caption{The details of BARA algorithm}
\label{alg:BARA_workflow}
\textbf{Input}: clients' local datasets $\left\{\mathcal{D}_{i=1} \right\}_{i}^{N}$; initial global model parameters $\omega^{0}$; test dataset $D_{test}$; GP prior $(\mu_{0}, \sigma_{0}, k)$; total reward budget $B_{total}$\\
\textbf{Output}: final global model parameters\\

\textbf{PS}:
\begin{algorithmic}[1]
% \STATE $\mathbf{n}_{0} \leftarrow \left\{ \right\}$\\
\STATE Test $\omega_{0}$ on $\mathcal{D}_{test}$ to get $a_{0}$\\
\STATE $B \leftarrow 0$\\
\FOR{$t=1$ to $T_{max}$}
\STATE Client $i$ report $b_{t,i}$ to PS, $i \le N$\\
% \tcc{Determine the number of participating clients}
\IF{$t<T_{0}$}
\STATE Randomly select $n_{t}$ from $\left\{ 1,2,\cdots,N-1 \right\}$\\
\ELSE
\STATE $n_{t} \leftarrow \argmax_{1 \le n<N}\mu_{t-1}(n)+ \sqrt{\beta_{t}}\sigma_{t-1}(n)$\\
\ENDIF
% \tcc{Client selection and reward allocation}
\STATE Sort all clients in descending order of $\frac{q_{t,i}}{b_{t,i}}$\\
\STATE $B_{t}(n_{t}) \leftarrow \sum_{i=1}^{n_{t}}\frac{b_{t,n_{t}+1}}{q_{t,n_{t}+1}}q_{t,i}$\\
\STATE $B \leftarrow B+B_{t}(n_{t})$\\
\IF{$B>B_{total}$}
\STATE Return $\omega_{t-1}$\\
\ENDIF
\FOR{$i=1$ to $n_t$ \textbf{in parallel}}
\STATE $r_{t,i} \leftarrow \frac{b_{t,n_{t}+1}}{q_{t,n_{t}+1}}q_{t,i}$\\
\STATE Send global model parameters $\omega_{t-1}$ to client $i$\\
\STATE $\nabla l(\omega_{t-1};\mathcal{D}_{i}) \leftarrow $ \textbf{ClientUpdate}($\omega_{t-1},\mathcal{D}_{i}$)\\
\ENDFOR

\STATE Perform model aggregation based on $\nabla l(\omega_{t-1};\mathcal{D}_{i})$\\
to obtain $\omega_{t}$ and test it on $\mathcal{D}_{test}$ to obtain $a_{t}$\\
% \STATE Approximate final model accuracy $a_{n_{t}}$ as in Eq. (\ref{con:cal_final_acc})\\
% \STATE Update $\mathbf{n}_t$ and $\hat{\mathbf{a}}_{t}$ with new observation\\
\STATE Update $\mathbf{M}_{t}$, $\widehat{\mathbf{M}}_{t}$ and perform Bayesian posterior update as in Eqs. \eqref{con:cal_mean} - \eqref{con:cal_variance} to obtain $\mu_{t}$ and $\sigma_{t}$\\
% \STATE Perform Bayesian posterior update as in Eqs. \eqref{con:cal_mean} - \eqref{con:cal_variance} to obtain $\mu_{t}$ and $\sigma_{t}$\\
\ENDFOR
\end{algorithmic}

\textbf{ClientUpdate($\omega_{t-1}, \mathcal{D}_{i}$)}:

\begin{algorithmic}[1]
\STATE Receive $\omega_{t-1}$ from the PS\\
\STATE Calculate $\nabla l(\omega_{t-1};\mathcal{D}_{i})$ based on $\omega_{t-1}$ and $\mathcal{D}_i$\\
\STATE Upload $\nabla l(\omega_{t-1};\mathcal{D}_{i})$ to the PS
\end{algorithmic}
\end{algorithm}

Until communication round $t$, we have made $t$ different choices of $n_t$. Thus, we can establish a GP with $t$ variables to predict the distribution of $z_{t+1}(n)$. To simplify our presentation, 
let $\mathbf{z}_t$ denote the vector of the first $t$ choices of $n$, i.e., $\mathbf{z}_t = \{z_1(n_1), z_2(n_2),\dots, z_t(n_t)\}$. 
Let $\mu(\mathbf{n}_{t}) =\{\mu(n_{1}), \mu(n_{2}), \dots, \mu(n_{t})\}$. The deviation between the observed value and the real value can be gauged by a zero-mean random noise $\epsilon_{t}\sim \mathcal{N}(0,\sigma^{2})$, which is independent with  time.
According to \cite{williams2006gaussian}, $(\mathbf{z}_t, z_{t+1}(n_{t+1}))$ is a sample drawn from  the following distribution:
%We assume the PS selects $n_{t+1}$ clients to participant in FL in the ($t+1$)-th communication round and obtain the new observation $a_{n_{t+1}}$, the joint distribution of $\hat{\mathbf{a}}_{t}$ and $a_{n_{t+1}}$ is given by:
\begin{align}
\label{EQ:Joint}
    % \begin{bmatrix}
    % \mathbf{z}_{t} \\
    % z_{t+1}(n_{t+1})
    % \end{bmatrix}\sim 
    \mathcal{N}\left ( 
    \begin{bmatrix}
    \mu(\mathbf{n}_{t}) \\
    \mu(n_{t+1})
    \end{bmatrix},
    \begin{bmatrix}
    \mathbf{K}_{t}+\sigma^{2}\mathbf{I} & \mathbf{k}_{t}(n_{t+1}) \\
    \mathbf{k}_{t}^{\top}(n_{t+1}) & k(n_{t+1},n_{t+1}) \\
    \end{bmatrix}
    \right ),
\end{align}
for $1 \le n_{t+1}<N$. Note that $n_{t+1}$ is the choice of round $t+1$, which has not occurred yet. 
Here $\mathbf{I}$ is a $t \times t$ identity matrix. $\mathbf{K}_{t}$ is the positive definite kernel matrix $[k(n,n')]_{\forall n,n' \in \mathbf{n}_{t}}$ and $\mathbf{k}_{t}(n_{t+1})=[k(n_{1},n_{t+1}),k(n_{2},n_{t+1}),\dots,k(n_{t},n_{t+1})]^{\top}$. It is easy to see that the joint distribution given in Eq.~\eqref{EQ:Joint} describes the relationship between $t+1$ variables. It can be regarded as the prior knowledge of Bayesian optimization to learn the posterior knowledge. At the end of communication round $t$, we can update the posterior knowledge, i.e.,  estimated mean and variance for $z_{t}(n)$ when choosing different $n$, as follows:
%The posterior distribution over objective function is a GP distribution again, with mean $\mu_{t}(n)$ and variance $\sigma_{t}^{2}(n)$:

\begin{align}
    \label{con:cal_mean}
    {\mu}_{t}(n)
    & = {\mathbf{k}}_{t}^{\top}(n){[{\mathbf{K}}_{t}+\sigma^{2}\mathbf{I}]}^{-1}\hat{\mathbf{a}}_{t},\\
    % k^{t}(n,n)
    % & = \mathbf{K}_{\ast \ast}^{t}(n)-{\widetilde{\mathbf{K}}_{\ast}^{t}(n)}^{\top}{[\widetilde{\mathbf{K}}^{t}+\sigma^{2}\mathbf{I}]}^{-1}\widetilde{\mathbf{K}}_{\ast}^{t}(n),\\
    \label{con:cal_variance}
    {\sigma}_{t}^{2}(n)
    & = k(n,n)-\
    {\mathbf{k}}_{t}^{\top}(n){[{\mathbf{K}}_{t}+\sigma^{2}\mathbf{I}]}^{-1}{\mathbf{k}}_{t}(n),
\end{align}
for $1 \le n<N$. Here, $\mathbf{\hat{a}}_t = \{ \hat{a}_0(n), \hat{a}_1(n), \dots, \hat{a}_{t}(n) \}$, which is computed based on the $n$-th column in matrix $\mathbf{\widehat{M}}_T$. In Eq.~\eqref{con:cal_mean}, we jointly utilize the results of Newton's polynomial interpolation, i.e., $\mathbf{\hat{a}}_t$,  and GP which captures the correlation when choosing different $n$ via the term ${\mathbf{k}}_{t}^{\top}(n){[{\mathbf{K}}_{t}+\sigma^{2}\mathbf{I}]}^{-1}$ to predict $z_t(n)$.  In this approach, we can fully utilize all historical records to make prediction.
In Eq.~\eqref{con:cal_variance}, the variance of $z_t(n)$ is updated accordingly to gauge the uncertainty of the estimation in Eq.~\eqref{con:cal_mean}. 

Note that $\mu_t(n) $ represents the expected final model accuracy by choosing $n$ clients per communication round predicted at communication round $t$. As $t$ increases, we will collect more and more information to continuously improve our prediction. For the $(t+1)$-th round, the decision aiming to maximize the final model accuracy should select the one that can maximize $\mu_t(n) $. However, considering the uncertainty of our prediction seized by $\sigma_t^2(n)$, it is better to add an exploration term based on $\sigma_t^2(n)$. Specifically, the decision of $n$ for communication round $t+1 $ is:
%Here the mean and variance are used to guide exploitation and exploration, respectively. Specifically, if the PS selects the number of participating clients with the largest mean value, it means that this choice will bring the highest benefits based on the historical observations. The choice with the largest variance means great uncertainty, which may bring higher benefits or lead to negative results. In our work, the PS considers balancing exploration and exploitation to make a decision:
\begin{align}
\label{EQ:SearchN}
    n_{t+1}=\argmax_{n: 1 \le n<N}\mu_{t}(n)+\sqrt{\beta_{t+1}}\sigma_{t}(n),
\end{align}
where $\sqrt{\beta_{t+1}}$ is a tuneable constant. $\sqrt{\beta_{t+1}}\sigma_{t}(n)$ is the exploration term, created based on previous empirical experience \cite{Srinivas2009GaussianPO}. The convergence property of Eq.~\eqref{EQ:SearchN} been proved in \cite{bogunovic2016time}, which can guarantee that searched $n_t$ will gradually approach to the optimal $n^*$. It will be further verified by our experiments in the next section. % gives the regret bound of Gaussian process upper confidence bound (GP-UCB) algorithm in the time-varying setting.

% An algorithm is to be no-regret if it satisfies $\displaystyle \lim_{t \to \infty} Reg_{t}/t=0$. The convergence property has been guaranteed in \cite{bogunovic2016time}, which gives the regret bound of Gaussian process upper confidence bound (GP-UCB) algorithm in the time-varying setting.

It is worth mentioning that a random strategy to select $n$ should be adopted at the  early stage of FL training because observation records are insufficient to establish the GP for accurately learning posterior knowledge. 
%process, the PS does not learn enough knowledge actually to make correct decisions. In order to ensure that the PS can learn real useful information instead of making decisions based on the very noisy observations. 
%We set a pure exploration stage to enrich prior knowledge before using Bayesian optimization to balance exploration and exploitation. 
Specifically, the online reward budget allocation algorithm has two stages: 1) a pure exploration stage and 2) an exploration-exploitation stage. 
In Stage 1, the PS randomly selects $n_t$ from $\left\{1,2,\cdots ,N-1 \right\}$ in each communication round and observations will be recorded. In Stage 2, the Bayesian optimization is performed with enriched prior knowledge. The first pure exploration stage can be executed for a fixed $T_{0}$ communication rounds.  Then, it will proceed to the exploration-exploitation stage in the remaining communication rounds. Note that Stage 1 should not exhaust the reward budget such that Stage 2 can be conducted. 

We describe the detailed procedure of BARA in Algorithm \ref{alg:BARA_workflow}. For each communication round $t<T_0$, the PS randomly selects $n_t$ from $\left\{1,2,\cdots ,N-1 \right\}$ in  Stage 1 (line 6). After client selection and model update (lines 10-20), the PS updates the matrices $\mathbf{M}_{t}$ and $\widehat{\mathbf{M}}_{t}$, which employed to Bayesian posterior update to obtain $\mu_{t}$ and $\sigma_{t}$ (line 22). In Stage 2, the PS balances exploration and exploitation based on the GP posterior (line 8). The time complexity of sorting all clients in descending order is $\mathcal{O}(N\log N)$ (line 10). Here $T = \max_{1 \le n<N} T(n)$. The time complexity of updating $\mathbf{M}_{t}$, $\widehat{\mathbf{M}}_{t}$ and Bayesian posterior update (line 22) are both $\mathcal{O}(T^{2})$. The overall time complexity of BARA is $\mathcal{O}(\max \left\{N\log N, T^{2} \right\})$, which is lightweight in comparison with training advanced FL models.
% {\bf YP: describe the line numbers of the main part contributed by you in Alg pseudo code. 
% Brielf analyze the computation complexity. 
% }
%Note that this is the joint probability distribution $P(\hat{\mathbf{a}}_{t},a_{n_{t+1}}|\mathbf{n}_{t},n_{t+1})$ over $\hat{\mathbf{a}}_{t}$ and $a_{n_{t+1}}$, we can derive the posterior distribution $P(a_{n_{t+1}}|\hat{\mathbf{a}}_{t},\mathbf{n}_{t},n_{t+1})$ over $a_{n_{t+1}}$ only from the joint probability distribution, which will be given later.

% Specifically, we use Ornstein-Uhlenbeck temporal covariance function to modify $\mathbf{K}_{t}$ as $\widetilde{\mathbf{K}}_{t}=\mathbf{K}_{t} \circ \mathbf{H}_{t}$ with $\mathbf{H}_{t}=[(1-\lambda)^{|i-j|/2}]_{i,j=1}^{t}$ and $\mathbf{k}_{t}(n)$ as $\widetilde{\mathbf{k}}_{t}(n)=\mathbf{k}_{t}(n) \circ \mathbf{h}_{t}$ with $\mathbf{h}_{t}=[(1-\lambda)^{(t+1-i)/2}]_{i=1}^{t}$. Given a fixed $\lambda$, the stale observations will be weighted less and less over time in Bayesian posterior update. And $\lambda$ controls how fast the weights of stale observations are dropped.

\begin{figure*}[!t]
    \vspace{-6mm}
    \centering
    \subfigure[MNIST]{
    \includegraphics[width=0.23\linewidth]{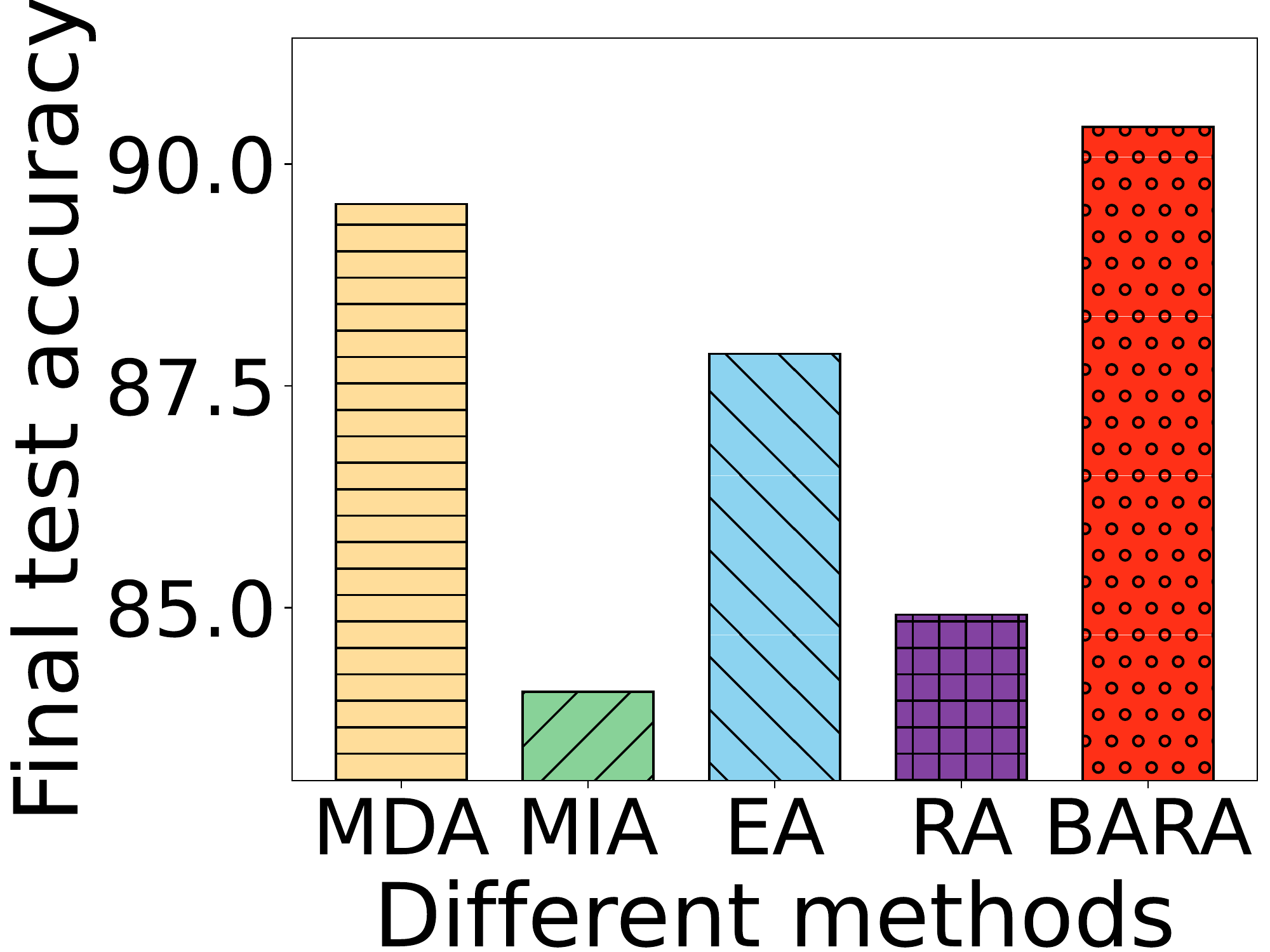}
    }
    \subfigure[FMNIST]{
    \includegraphics[width=0.23\linewidth]{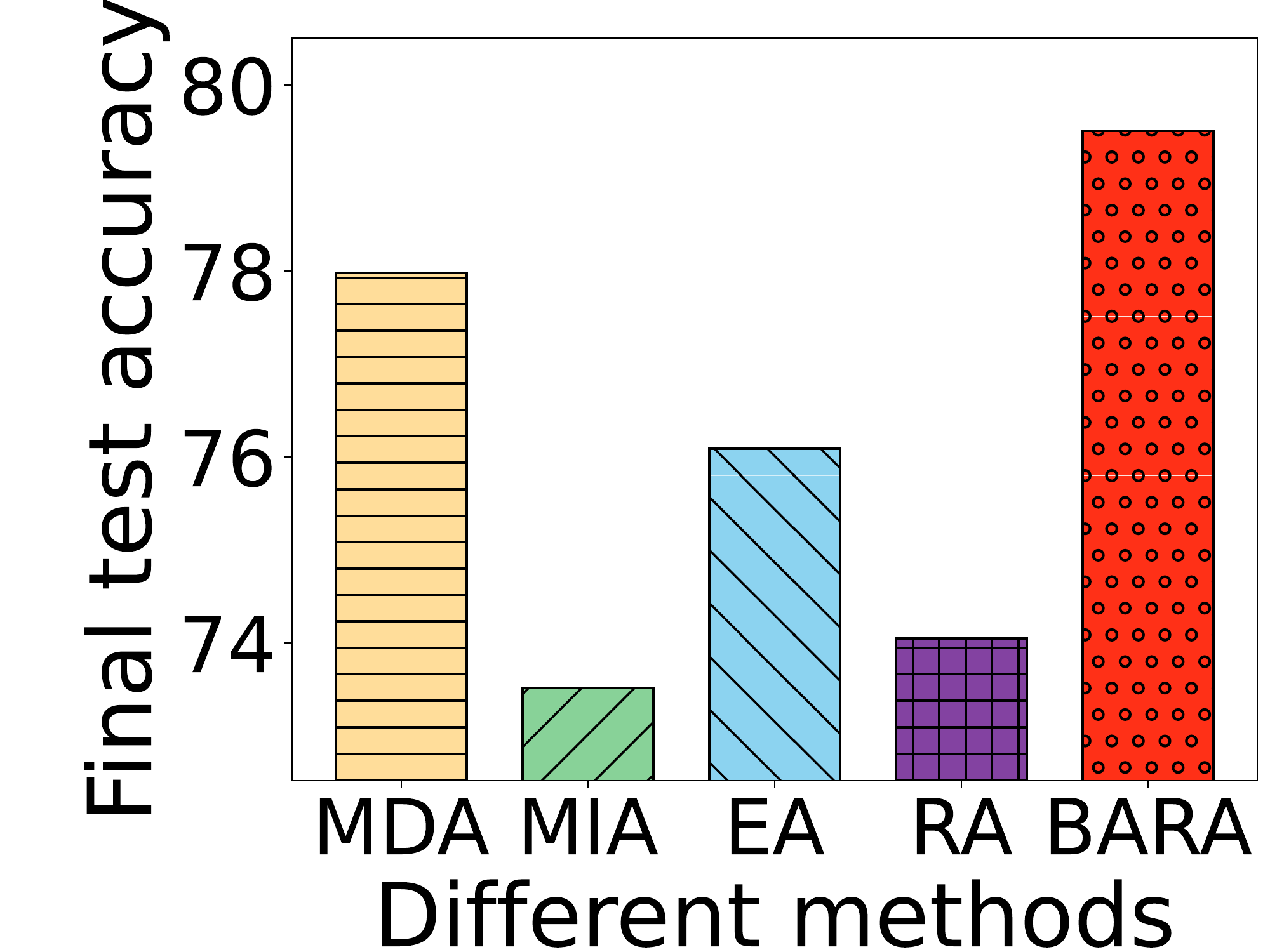}
    }
    \subfigure[CIFAR-10]{
    \includegraphics[width=0.23\linewidth]{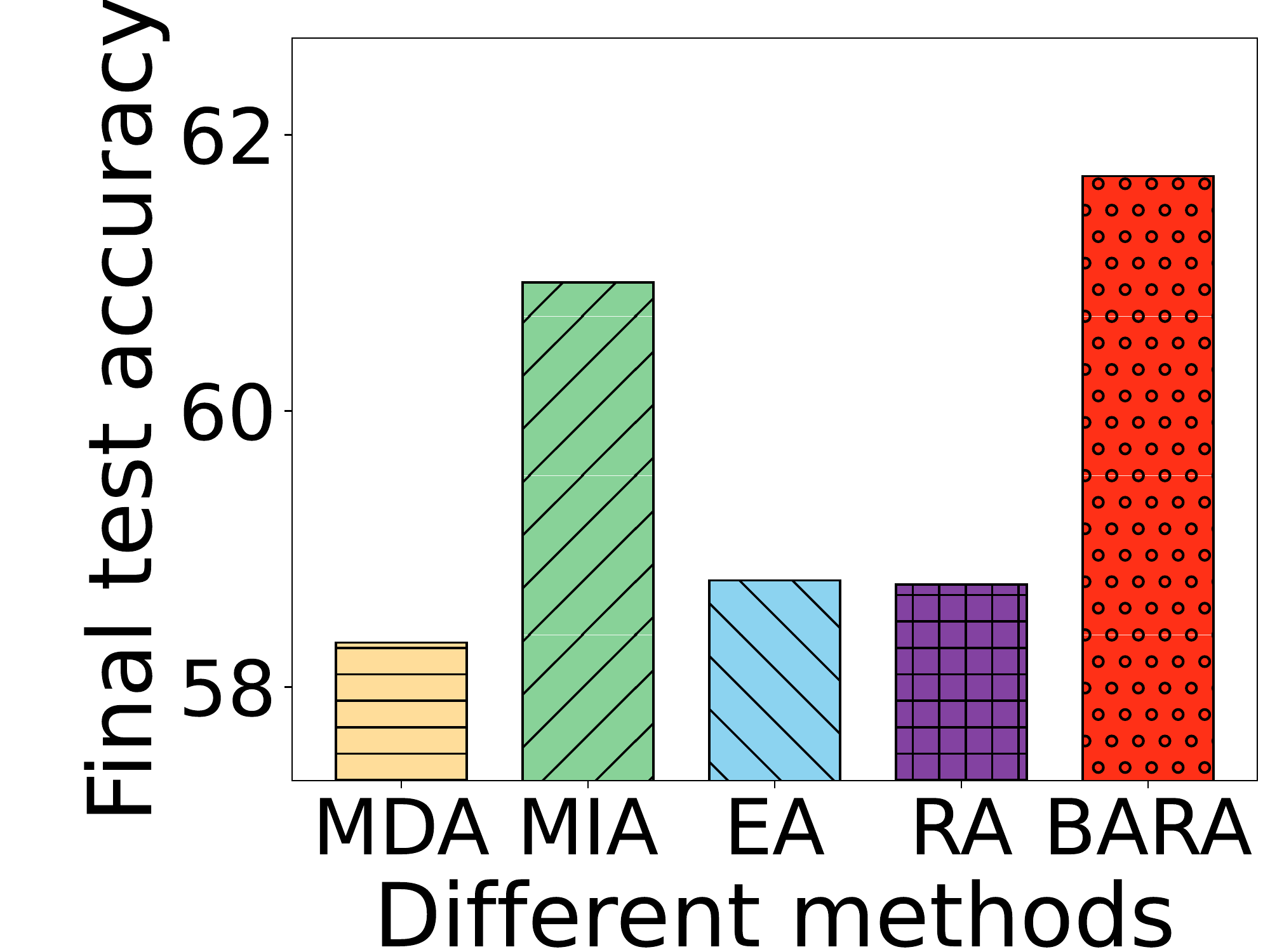}
    }
    \subfigure[CIFAR-100]{
    \includegraphics[width=0.23\linewidth]{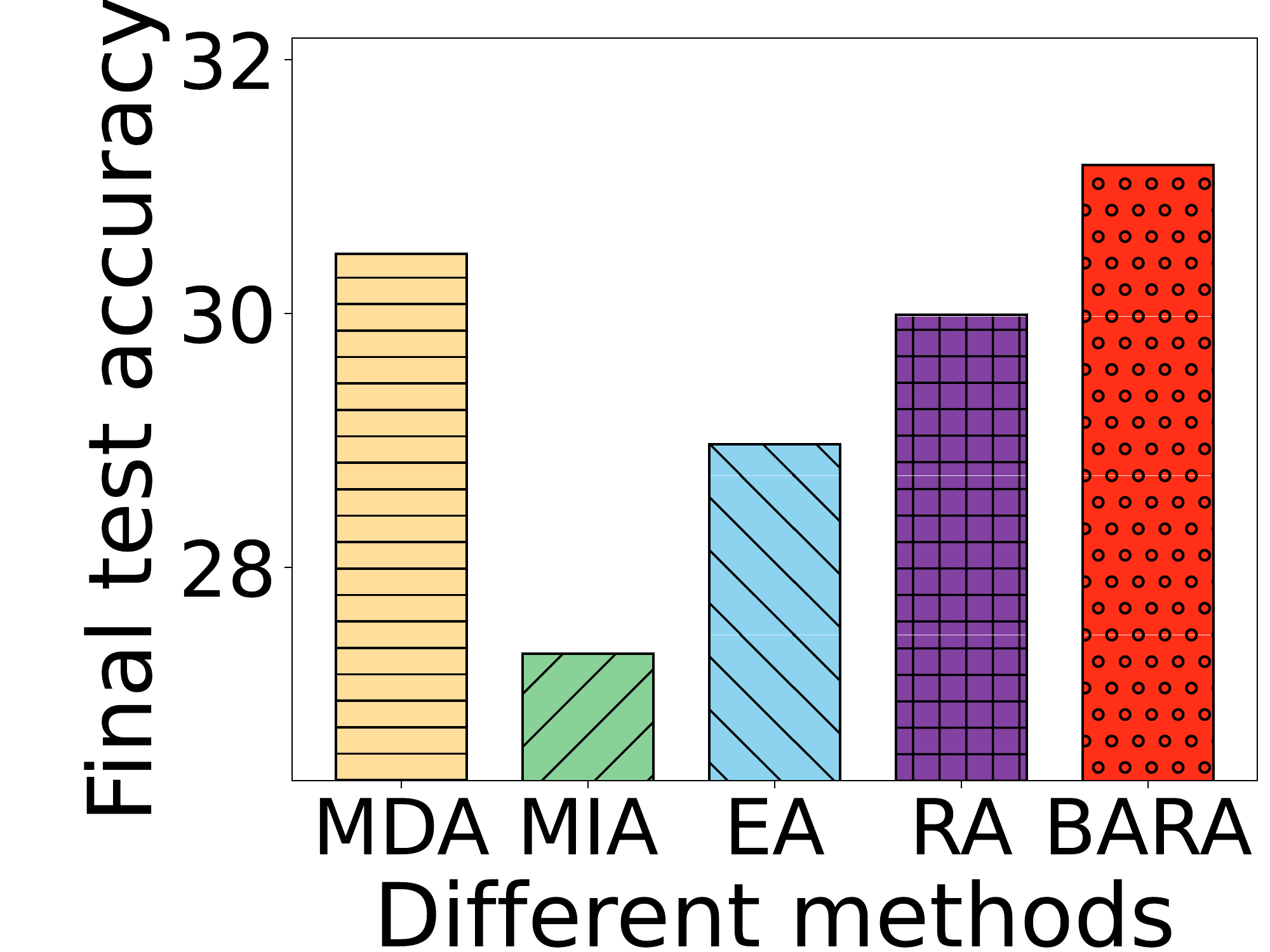}
    }
    \vspace{-3.5mm}
    \caption{Comparing final model accuracy of different reward budget allocation methods on four datasets.}
    \label{fig:diff_allocation_methods}
    % \vspace{-3.5mm}
\end{figure*}

\begin{figure*}[!t]
    \vspace{-5mm}
    \centering
    \subfigure[MNIST]{
    \includegraphics[width=0.23\linewidth]{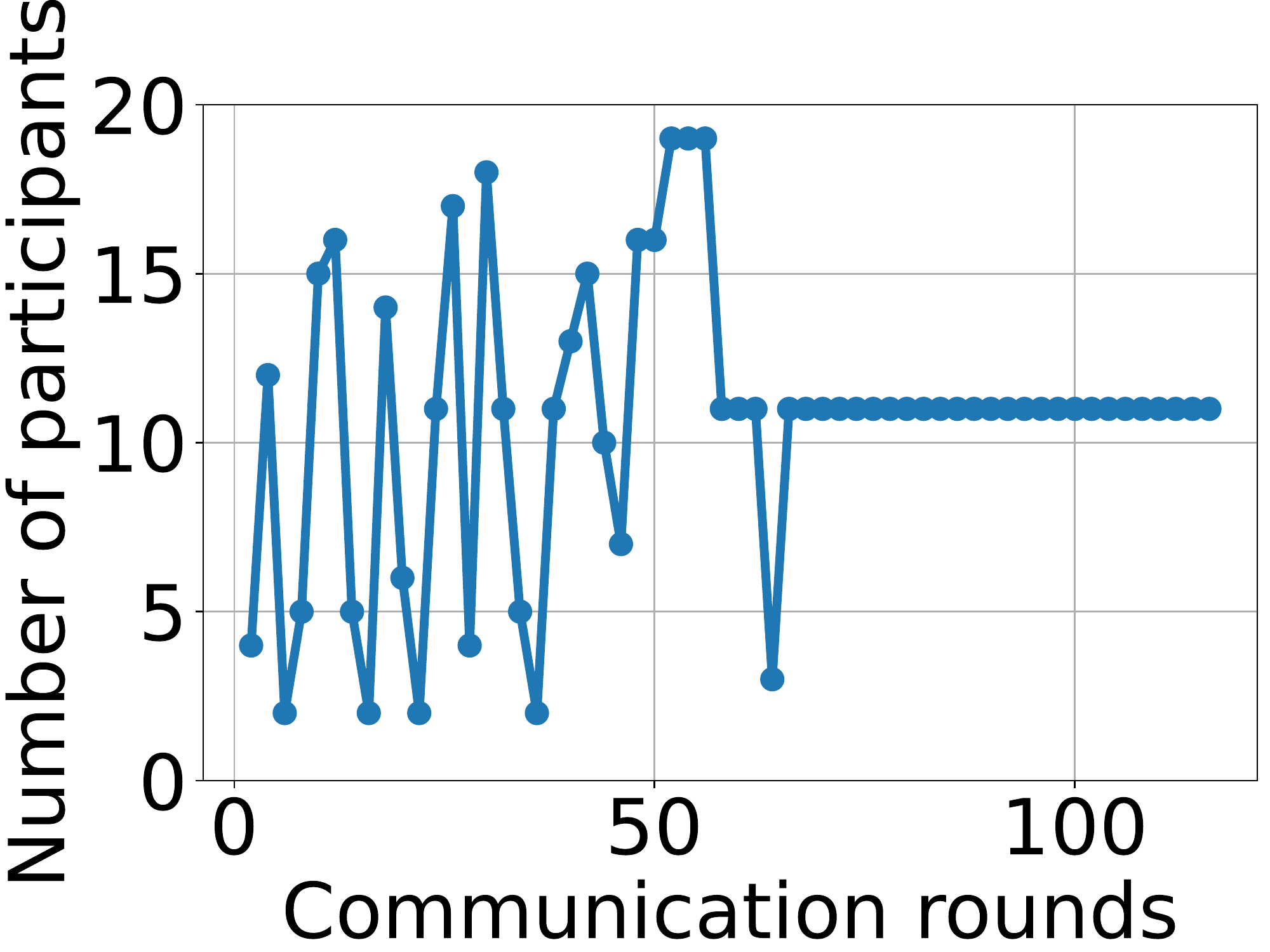}
    }
    \subfigure[FMNIST]{
    \includegraphics[width=0.23\linewidth]{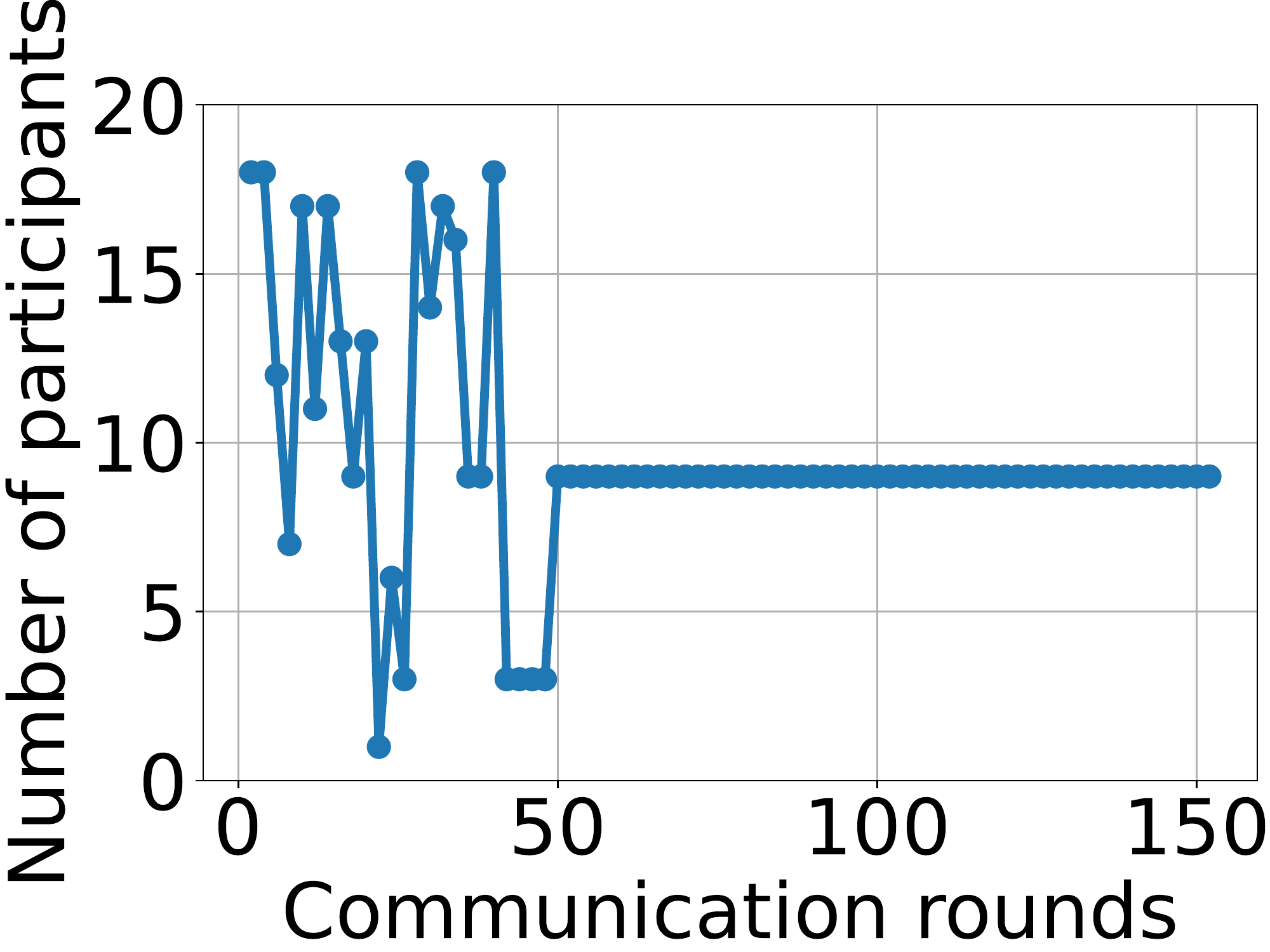}
    }
    \subfigure[CIFAR-10]{
    \includegraphics[width=0.23\linewidth]{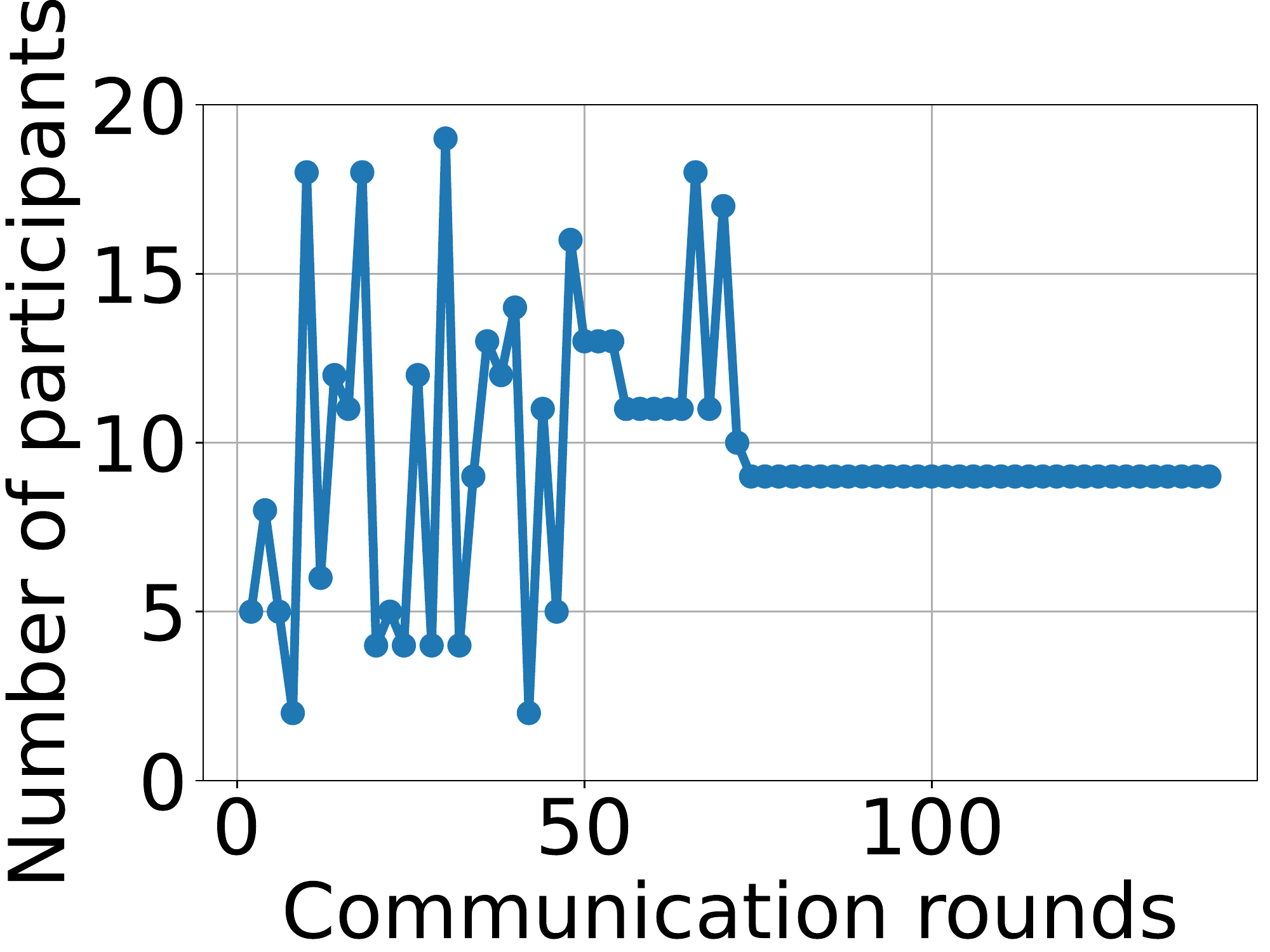}
    }
    \subfigure[CIFAR-100]{
    \includegraphics[width=0.23\linewidth]{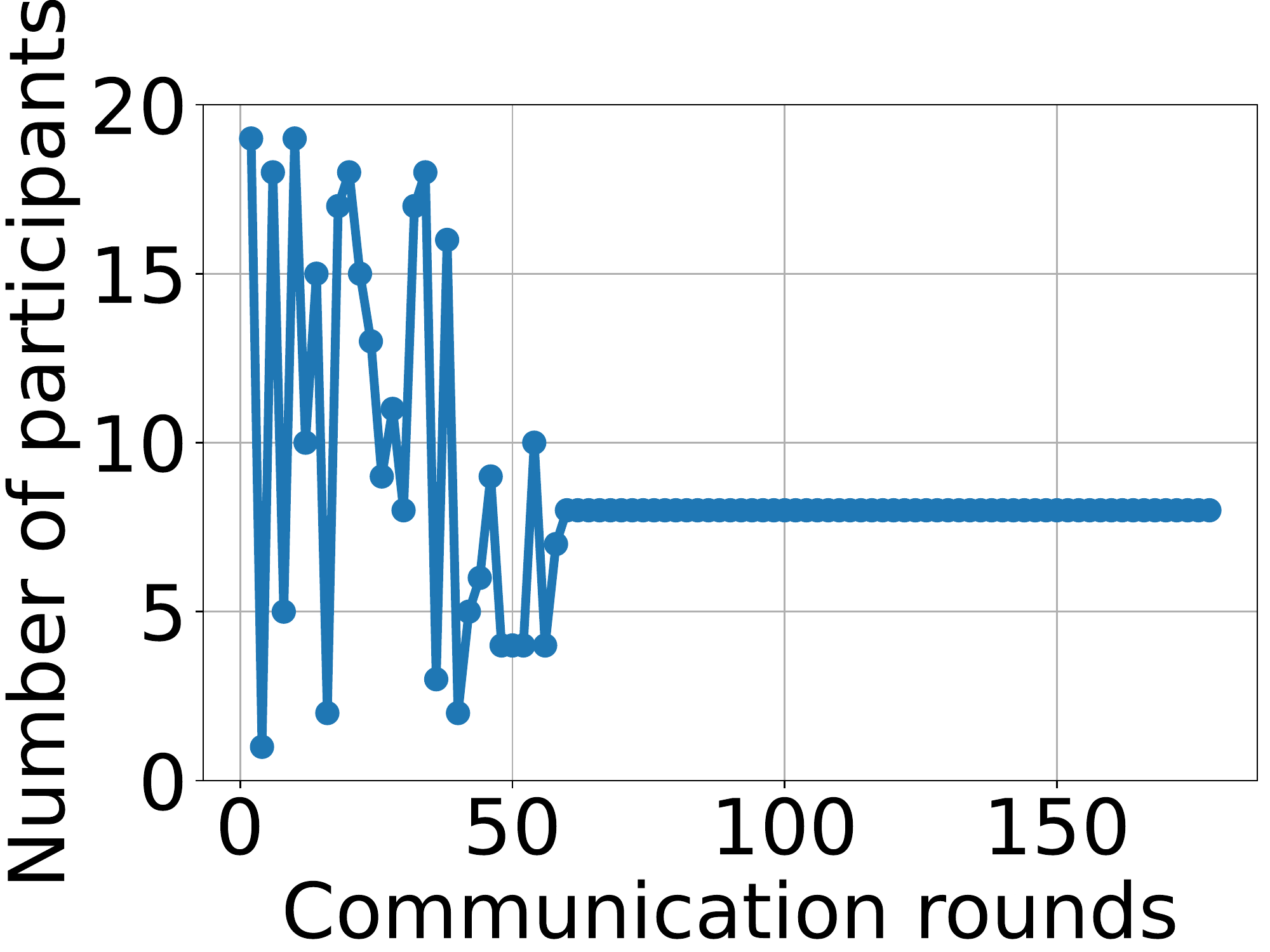}
    }
    \vspace{-3.5mm}
    \caption{The number of participants per communication round selected by BARA on four datasets.}
    \label{fig:participant_number}
    % \vspace{-3.5mm}
\end{figure*}

\begin{figure*}[!t]
    \vspace{-5mm}
    \centering
    \subfigure[MNIST]{
    \includegraphics[width=0.23\linewidth]{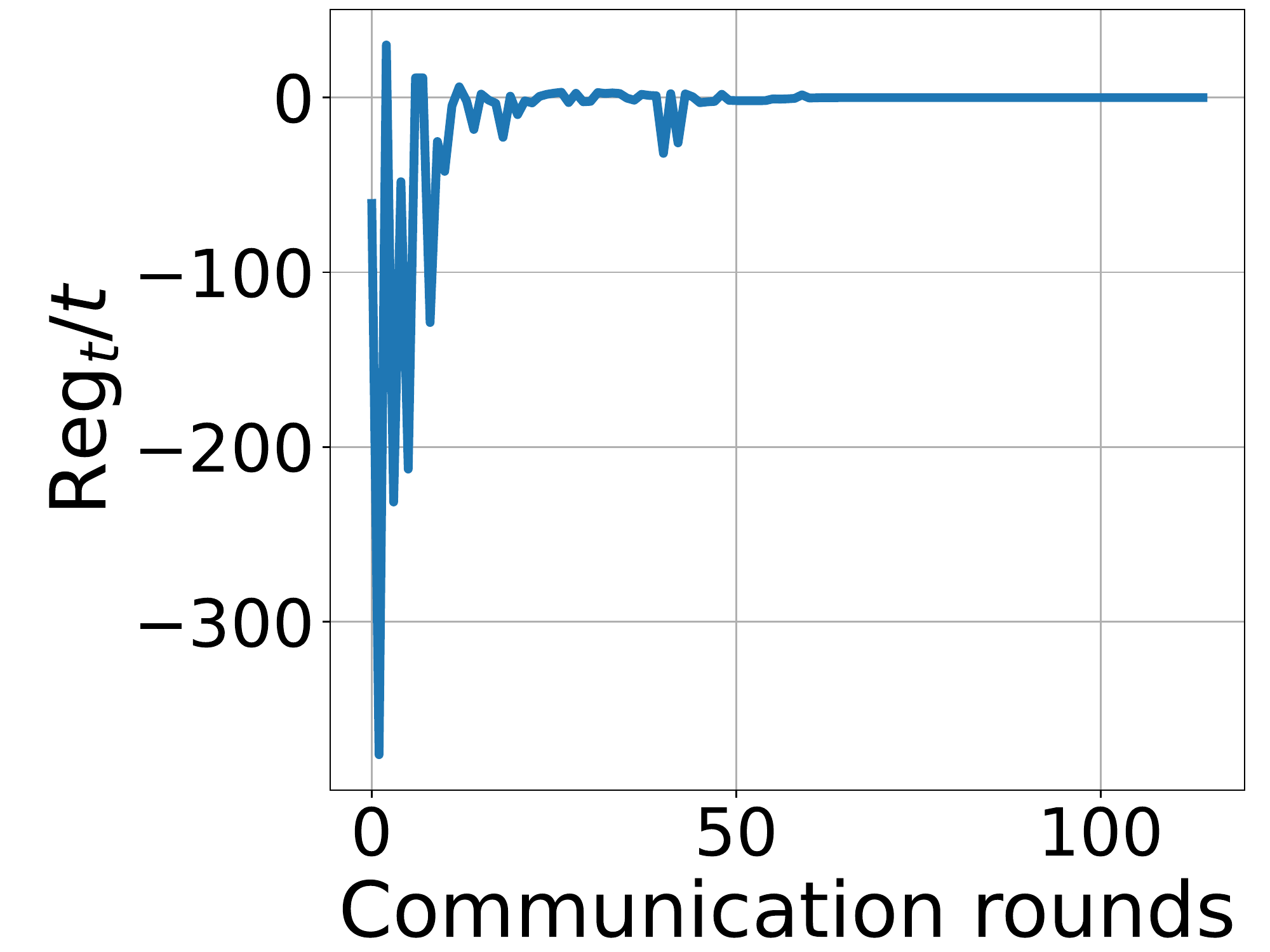}
    }
    \subfigure[FMNIST]{
    \includegraphics[width=0.23\linewidth]{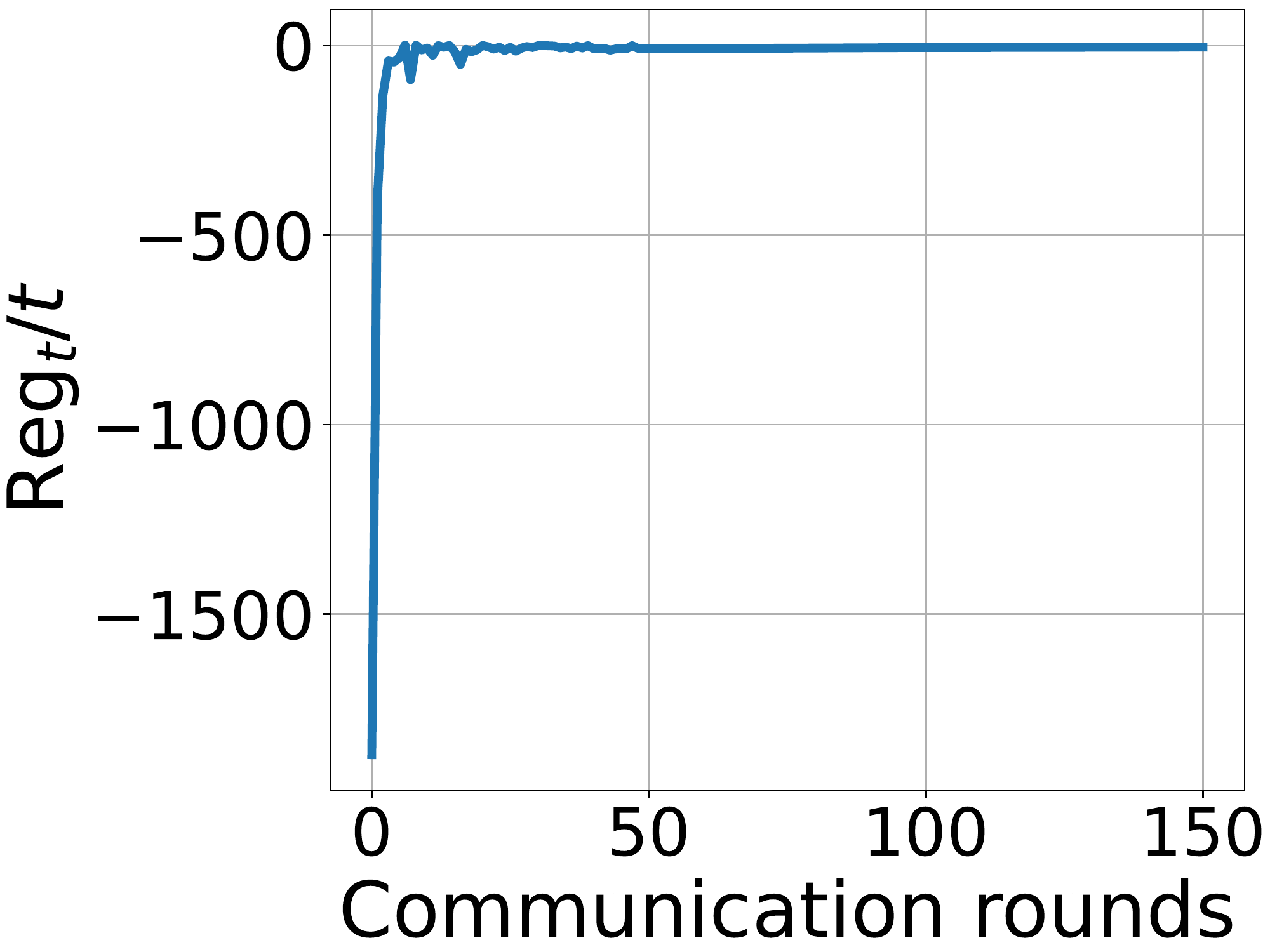}
    }
    \subfigure[CIFAR-10]{
    \includegraphics[width=0.23\linewidth]{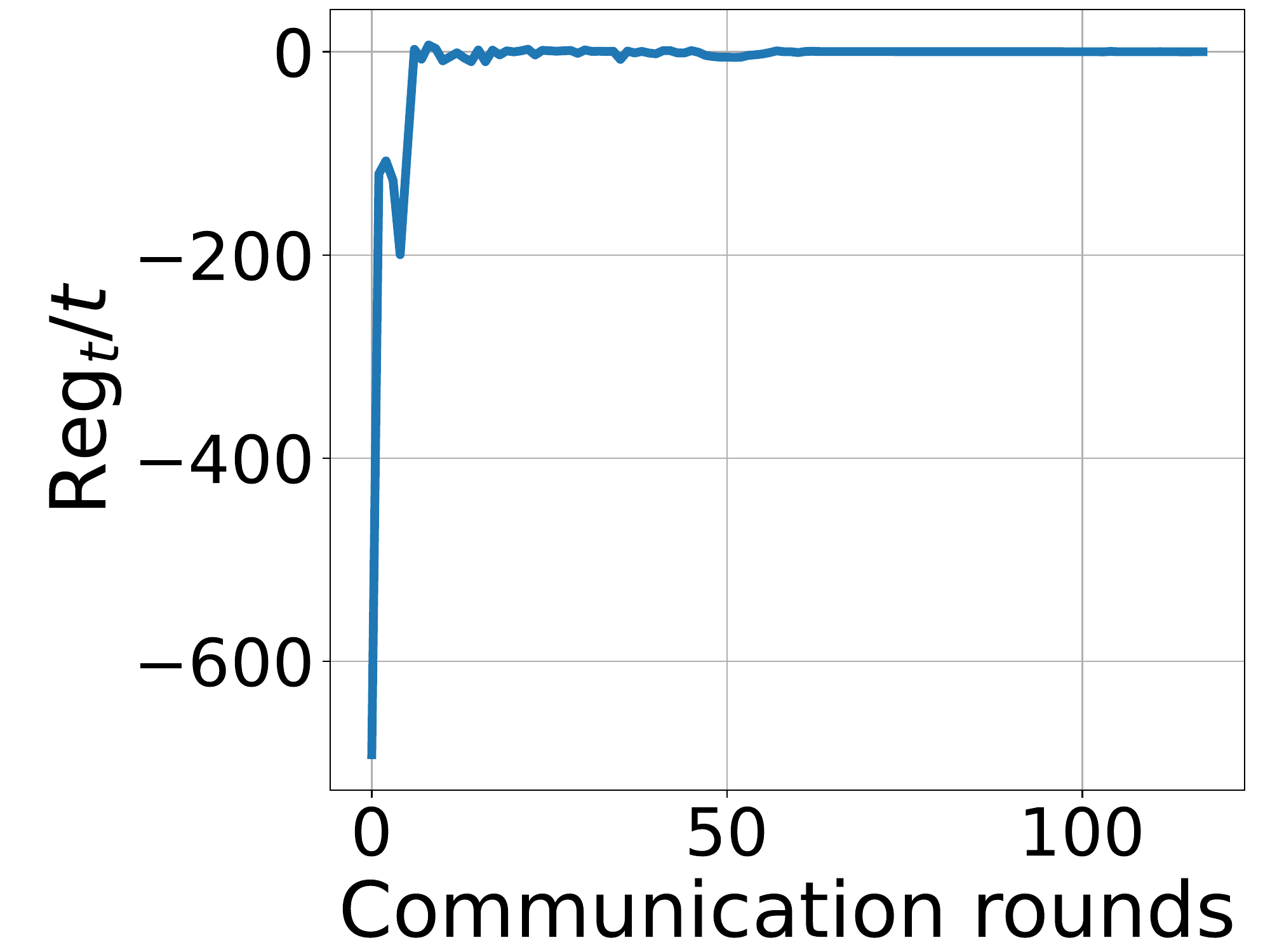}
    }
    \subfigure[CIFAR-100]{
    \includegraphics[width=0.23\linewidth]{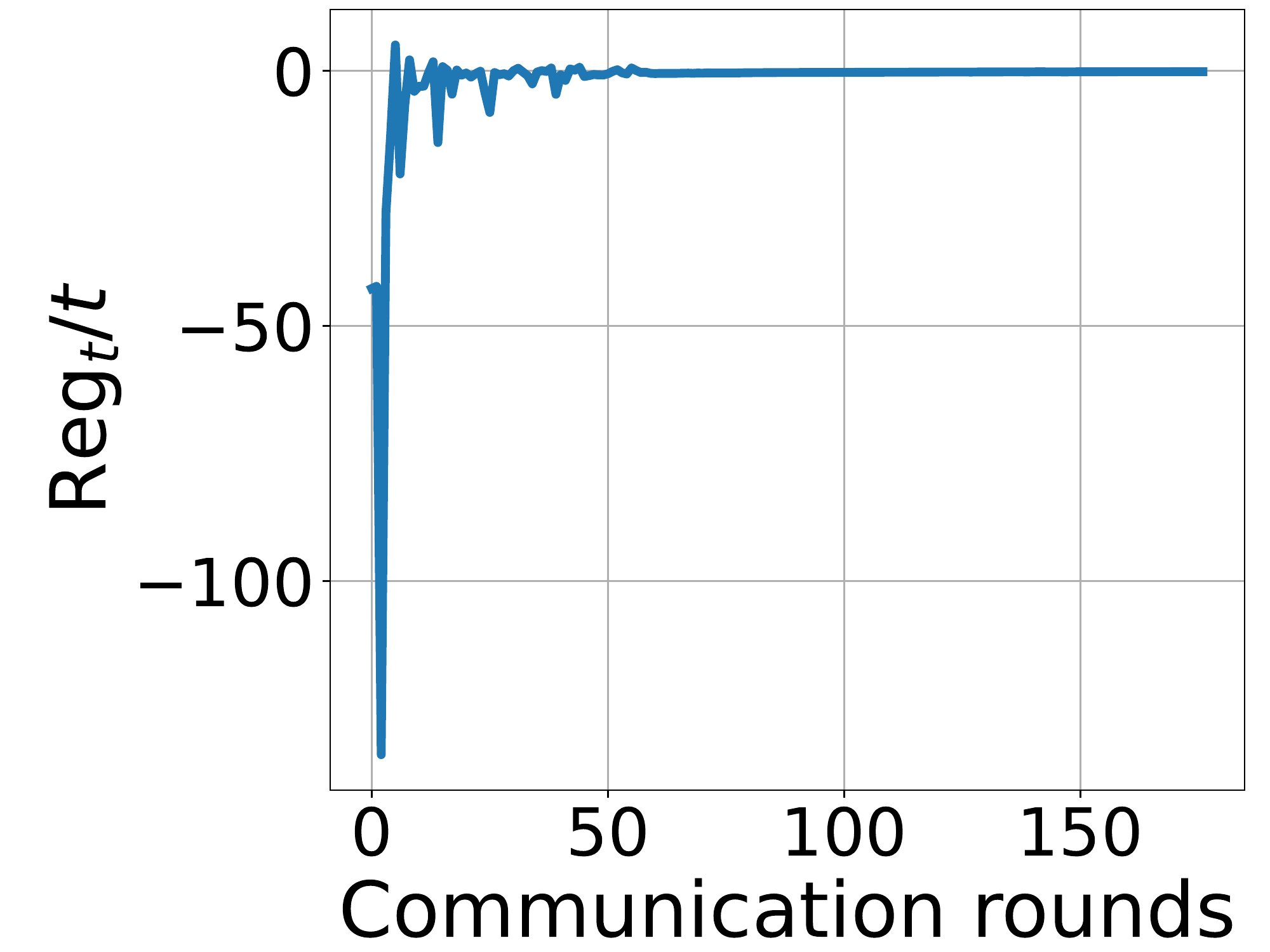}
    }
    \vspace{-3.5mm}
    \caption{The regret  (showing the gap to the optimal value) per communication round of BARA.}
    \label{fig:regret_time}
    \vspace{-4mm}
\end{figure*}

\section{Experiments}
\label{sec:experiments}

% By using benchmark datasets, we evaluate our algorithm in comparison with baseline in this section. 
\subsection{Experimental Setups}

\subsubsection{Datasets and Models}

We use four public datasets for experiments: MNIST \cite{lecun1998gradient}, Fashion-MNIST (also abbreviated as FMNIST) \cite{xiao2017fashion}, CIFAR-10 and CIFAR-100 \cite{krizhevsky2009learning} datasets. Similar to \cite{McMahan2016CommunicationEfficientLO}, we train a multilayer perceptron (MLP) model that consists of 2-hidden layers for classifying the MNIST dataset. A CNN (convolutional neural network) model that consists of two convolution layers (each followed by a max pooling layer and ReLU activation), then followed by a fully connected layer is trained for  classifying the FMNIST dataset. For CIFAR-10 and CIFAR-100 datasets, we train the CNN model with the same structure as that in  \cite{mills2021multi},
which  consists of two convolutional pooling layers, two batch normalization layers and two fully connected layers.

In real scenarios, the typical data distribution on FL suffers from statistical heterogeneity due to the fact that the training data owned by a particular client is usually related with user-specific features. Therefore, we allocate training datasets to clients in a non-IID manner. According to \cite{Chen2022PersonalizedFL}, for each dataset, we first sort samples by their labels and then split them into $2N$ shards equally. Each of $N$ clients randomly selects 2 shards.
% {\bf YP: you emphasize the data size varies widely, but in your setting, the size of every client data is identical? }

% We use three public datasets for experiments: MNIST \cite{lecun1998gradient}, Fashion-MNIST (also abbreviated as FMNIST) \cite{xiao2017fashion}, CIFAR-10 and CIFAR-100 \cite{krizhevsky2009learning} datasets. MNIST is a dataset of handwritten digit pictures, in which each sample is a grayscale picture with a length of 28 and a width of 28.  The content of each picture is a 0-9 digit. FMNIST covers a total of 70,000 different product front images from 10 categories. CIFAR-10 is a dataset with 10 classes, and each sample is an RGB color image. For different datasets, we train different models for classifying these images with correct labels. We employ a MLP (multilayer perceptron) model that consists of three fully connected layers for processing MNIST and FMNIST datasets, a CNN (convolutional neural network) model that consists of two convolution layers (each followed by a max pooling
% layer and ReLU activation), then followed by three fully
% connected layers for processing CIFAR-10.

\subsubsection{Parameter Settings}

We set the total number of clients $N$ as 20. Based on the empirical results in \cite{deng2021fair}, we set the  maximum number   of communication rounds, i.e., $T_{max}$, and the total reward budget of the PS, i.e., $B_{total}$, as $200$ and $1,500$, respectively. We implement a typical reverse auction-based incentive mechanism for FL: bid price first (i.e., all clients are of equal quality) \cite{deng2021fair}. Each client's bid for participating in each FL training round is independently sampled from a uniform distribution $U(0.5, 1.5)$. Referring to \cite{bogunovic2016time}, we set hyperparameters in the GP as  $\sqrt{\beta_{t}}=0.8\log(0.4t)$, the length scale parameter $l = 0.2$ for squared exponential kernel and $\lambda$ as 0.001. We set the noise variance $\sigma^{2}$ to 0.01. The pure exploration stage runs for initial $T_{0}=40$ communication rounds.

% According to prior work \cite{zheng2021incentive},  

% \subsubsection{Data Partitioning}

% \begin{figure*}[h]
%     \vspace{-4.5mm}
%     \centering
%     \subfloat[MNIST]{
%     \includegraphics[width=0.24\linewidth]{experiments/mnist_ours_fair.eps}
%     }\hfill
%     \subfloat[FMNIST]{
%     \includegraphics[width=0.24\linewidth]{experiments/fmnist_ours_fair.eps}
%     }\hfill
%     \subfloat[CIFAR-10]{
%     \includegraphics[width=0.24\linewidth]{experiments/cifar10_ours_fair.eps}
%     }\hfill
%     \subfloat[CIFAR-100]{
%     \includegraphics[width=0.24\linewidth]{experiments/cifar10_ours_fair.eps}
%     }
%     \vspace{-3mm}
%     \caption{The impact of BARA on FAIR}
%     \label{fig:case_study_fair}
%     % \vspace{-3.5mm}
% \end{figure*}

% \begin{figure*}[h]
%     \vspace{-4.5mm}
%     \centering
%     \subfloat[MNIST]{
%     \includegraphics[width=0.24\linewidth]{experiments/mnist_ours_fl_market.eps}
%     }\hfill
%     \subfloat[FMNIST]{
%     \includegraphics[width=0.24\linewidth]{experiments/fmnist_ours_fl_market.eps}
%     }\hfill
%     \subfloat[CIFAR-10]{
%     \includegraphics[width=0.24\linewidth]{experiments/cifar10_ours_fl_market.eps}
%     }\hfill
%     \subfloat[CIFAR-100]{
%     \includegraphics[width=0.24\linewidth]{experiments/cifar10_ours_fl_market.eps}
%     }
%     \vspace{-3mm}
%     \caption{The impact of BARA on FL-Market}
%     \label{fig:case_study_fl_market}
%     % \vspace{-3.5mm}
% \end{figure*}

\subsubsection{Compared Baselines}

%To  evaluate the performance of our design, 
We compare BARA with the following reward budget allocation baseline methods:
\begin{itemize}
    \item \textbf{Even allocation (EA):} The PS allocates the total reward budget $B_{total}$ evenly to each communication round (i.e., $B_{t}=\frac{B_{total}}{T_{max}}$), which is commonly adopted in existing works \cite{deng2021fair,zhang2021incentive,zheng2021incentive}.
    % {\bf YP: cite some papers here}
    \item \textbf{Monotonically increasing allocation (MIA):} The allocated reward budget $B_{t}$ for round $t$ is  a monotonically increasing function  with  $t$ (i.e., $B_{t}=2\frac{B_{total}}{T_{max}^{2}}t$).
    \item \textbf{Monotonically decreasing allocation (MDA):} The allocated reward budget $B_{t}$ of round $t$ is a monotonically decreasing function with $t$ (i.e., $B_{t}=-2\frac{B_{total}}{T_{max}^{2}}t+2\frac{B_{total}}{T_{max}}$).
    \item \textbf{Random allocation (RA):} In each communication round, the PS randomly selects the number of participating clients from $\left\{1,2,\cdots ,N-1 \right\}$. The FL training process halts once the total reward budget is used up.
\end{itemize}
% {\bf YP: can you cite some paper to support each baseline?
% }

\subsubsection{Evaluation Metrics}
We adopt two metrics, \emph{test accuracy} and \emph{regret}, to evaluate our algorithm. 
%demonstrate the effectiveness of our algorithm. 
Test accuracy evaluates the accuracy of $\omega^{t}$ on $\mathcal{D}_{test}$ in each communication round $t$. By comparing test accuracy, we can evaluate how much performance gain can be achieved by optimizing the reward budget allocation in FL. 
%we can calculate the test accuracy $a^{t}$  and higher test accuracy means better algorithm's performance. 
Regret evaluates the gap between the solution of our algorithm and the theoretically optimal solution. To obtain the theoretically optimal solution, we enumerate $n$ in FL to find which $n*$ can achieve the highest final model accuracy on the test set. Due to the  limited reward budget in practice, it is impossible to enumerate all possible $n$. Thus, the constraint of the reward budget is not considered for searching $n^*$. 
%Moreover, we denote the final test accuracy with the optimal number of participating clients as $a_{T(n^{\ast})}(n^{\ast})$. In each communication round, we define the regret as the difference between the final model accuracy with the optimal action $n^{\ast}$ and that of the observed final model accuracy with the actual selected action $n_{\tau}$. After $t$ communication rounds, 
Once $n^*$ is determined, we define the regret at round $t$ as $Reg_{t}=\sum_{\tau=1}^{t}(a_{T(n^{\ast})}(n^{\ast})-\hat{a}_{\tau}(n_{\tau}))$ where $\hat{a}_{\tau}(n_{\tau})$ is the estimation of model accuracy predicted by our model. % in Sec~\ref{sec:methodology}. 
Intuitively, if $Reg_{t}$ approaches $0$ with $t$, i.e., $\displaystyle \lim_{t \to \infty} Reg_{t}/t=0$, it implies % An algorithm is to be no-regret if it satisfies , which implies 
that BARA can approximately find the optimal solution after a certain number of rounds. 
%on average the PS performs almost as well as the optimal fixed action in hindsight.

% The convergence property has been guaranteed in \cite{bogunovic2016time}, which gives the regret bound of Gaussian process upper confidence bound (GP-UCB) algorithm in the time-varying setting.
% Since the squared exponential kernel we used satisfies the smoothness assumptions in \cite{bogunovic2016time}, the convergence property of Gaussian process upper confidence bound (GP-UCB) algorithm in the time-varying setting is guaranteed. Defining $C_{1}=8/\log(1+\sigma^{-2})$, the regret bound is:
% \begin{align}
%     Reg_{t} 
%     & \le \sqrt{C_{1}t\beta_{t}\tilde{\gamma}_{t}}+2\\
%     \label{con:regret_bound}
%     & \le \sqrt{C_{1}t\beta_{t}
%     \left ( \frac{t}{\tilde{t}}+1 \right )
%     \left ( \tilde{\gamma}_{\tilde{t}}+\tilde{t}^{3}\epsilon \right )
%     +2}
% \end{align}
% with probability at least $1-\delta$, where \ref{con:regret_bound} holds for any $\tilde{t} \in \left\{1,2,\cdots ,t \right\}$. The detailed proof and the expression of $\beta_{t}$ are given in \cite{bogunovic2016time}.

\subsection{Experimental Results}

% \subsubsection{Evaluating Reward Budget Allocation Methods}

We first conduct experiments to compare BARA with other reward budget allocation methods. The experimental results are plotted in Figure \ref{fig:diff_allocation_methods} with the x-axis representing different reward budget allocation methods and the y-axis representing the final model accuracy on $\mathcal{D}_{test}$. The results in Figure \ref{fig:diff_allocation_methods} manifest that BARA can significantly outperform other baselines in term of final model accuracy. 
The analysis of experimental results are presented in Appendix A.

% \ref{appendix_experiment_analysis}

%For MDA there are many clients are involved in FL in the early stage of training. However, the total reward budget will be exhausted quickly, and the number of participating clients will decrease over time in MDA, which makes it difficult to improve the performance further. The model performance in MIA improves greatly over time, but the model cannot be effectively trained in the early stage of training due to the small number of participating clients.
% show that: 1) BARA outperforms other baselines both in terms of convergence speed and final model accuracy. 2) The reason why EA and MDA converge fast is that many clients are involved in FL in the early stage of training. However, the total reward budget will be exhausted quickly, and the number of participating clients will decrease over time in MDA, which makes it difficult to improve the performance further. 3) The model performance in EA and MIA improves greatly over time, but the model cannot be effectively trained in the early stage of training due to the small number of participating clients.

Next, we investigate the learning process of BARA for searching the optimal number of participating clients. Based on experimental results in Figure~\ref{fig:diff_allocation_methods}, we plot the number of participating clients selected by BARA for each dataset in  Figure \ref{fig:participant_number}. Here,  the x-axis represents the communication round and the y-axis represents the number of participating clients per communication round. From Figure \ref{fig:participant_number}, we can observe that 1) In the initial $T_0=40$ communication rounds, the number of participating clients fluctuate over time as the PS 
randomly selects $n$ for participating in FL. 2) The number of participating clients quickly converges to a stable value beyond the critical point $T_0=40$  indicating that BARA can efficiently explore $n^*$ with sufficient historical records. 

%tries various numbers of participating clients in the pure exploration stage to enrich the prior knowledge. Then BARA converges to the optimal solution after a short period of instability in the exploration and exploitation stage. 

To further verify the effectiveness of BARA, we evaluate the regret of BARA for each dataset. In each communication round, we plot $Reg_{t}/t$ in Figure \ref{fig:regret_time} with the x-axis representing the communication round and y-axis representing $Reg_{t}/t$. From Figure \ref{fig:regret_time},  we can observe the fast convergence of the regret curve  when $t>T_0$. As the regret approaches to $0$,  it  implies that BARA finds $n^*$ for determining the number of participating clients. 

BARA is applicable for various different incentive mechanisms. To demonestrate this generic value of BARA, we implement two more typical reverse auction-based incentive mechanisms for FL. Their performance can be further enhanced by incorporating BARA into their mechanisms.
The detailed experimental results are presented at Appendix B.

% \ref{appendix_case_study}
% {\bf YP: not sure why we need to plot these small subfigures? The font is to small to read. }

%a continuous decrease of the regret per communication round, which implies BARA can gradually find the optimal number of participating clients.

% {\bf YP: which mechanism you actually used for experiments in Fig 2,3 4? is it FAIR or another one?? should state it clearly. If you use FAIR for figs 2,3,4, not sure whether it is necessary to have table 1.  }
%Thewith x-axis representing the communication round and y-axis representing model accuracy. Table \ref{tab:case_study_fair} and Table \ref{tab:case_study_fl_market} show that applying BARA to FAIR and FL-Market can greatly improve their final model accuracy since BARA finds the optimal number of participating clients by studying the specific model training process. Therefore, BARA can make corresponding optimal decisions even faced with different scenarios.

\section{Conclusion}
\label{sec:conclusion}

To our best knowledge, our work is the first one to  investigate the reward budget allocation problem between training rounds in federated learning given a limited total budget. 
Due to the complicated relationship between reward budget allocation and final model utility, we established a Gaussian process (GP) model to predict model utility with respect to reward budget allocation. To expand the historical knowledge for building the GP model, Newton's polynomial interpolation was applied to generate artificial records. We further employed the Bayesian optimization to determine the reward budget allocation to maximize the predicted final model utility. Based on our analysis, an online reward budget allocation algorithm called BARA was proposed, which is lightweight for implementation. %We design an approximating final model accuracy algorithm using Newton's polynomial interpolation. After that, BARA separates Bayesian optimization into two stages for enriching prior knowledge and balancing exploration-exploitation, respectively. 
Finally, extensive experiments were conducted to demonstrate the effectiveness of BARA by extensively improving model accuracy compared with baselines. 
%In our future work, we will extend the application of our approach by  incorporating it into various incentive mechanisms in FL. 

\section*{Acknowledgement}
This work was supported by the National Natural Science Foundation of China under Grants U1911201, U2001209, 62072486, and the Natural Science Foundation of Guangdong Province under Grant 2021A1515011369.

% \newpage

%% The file named.bst is a bibliography style file for BibTeX 0.99c
\bibliographystyle{named}
\bibliography{ijcai23}

\newpage
\clearpage

\appendices

\section{Analysis of Experimental Results}
\label{appendix_experiment_analysis}

\begin{figure}[h]
    \centering
    \includegraphics[width=0.6\linewidth]{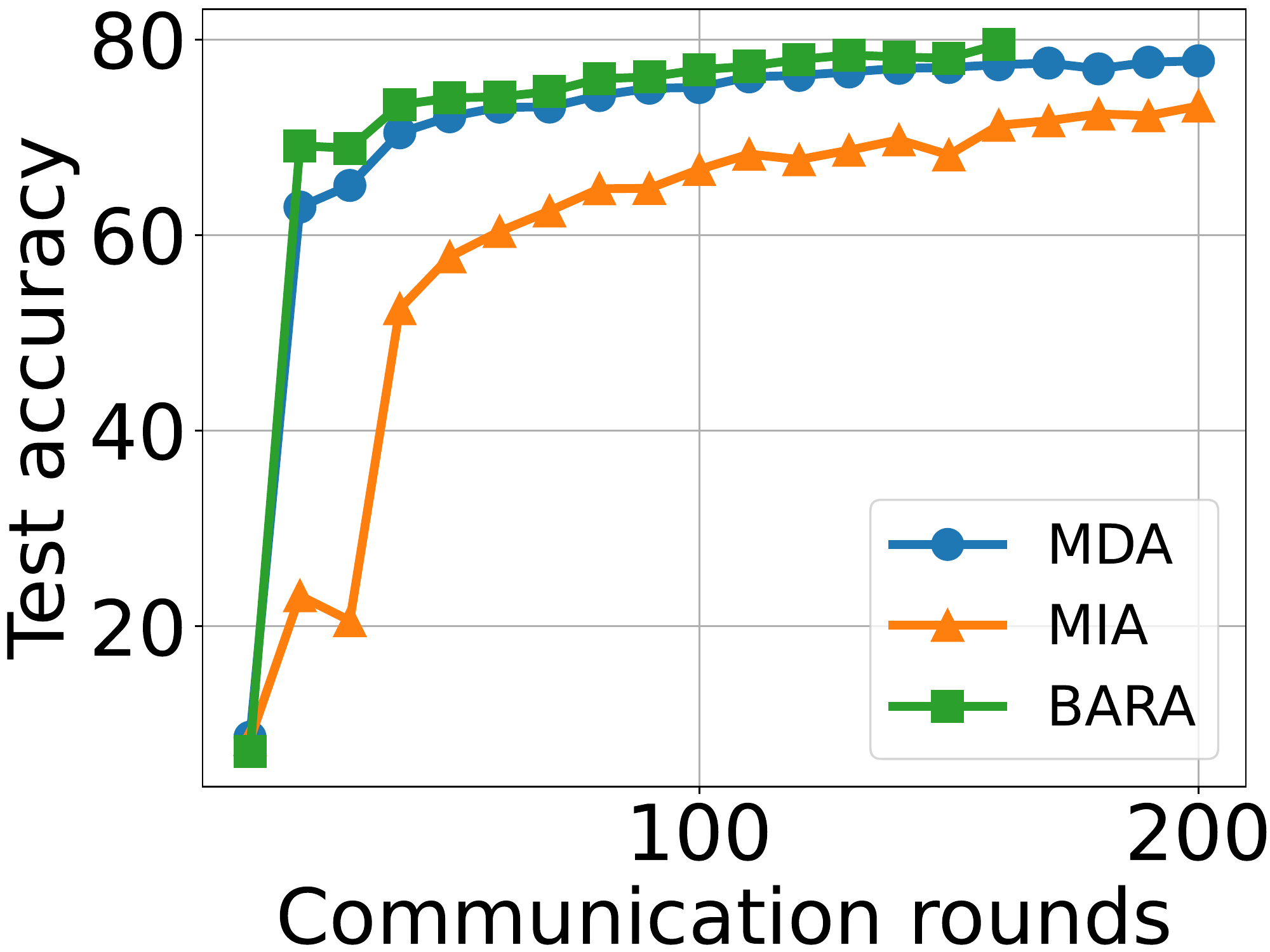}
    % %\vspace{-3mm}
    \caption{Test accuracy of different reward budget allocation methods on FMNIST dataset.}
    \label{fig:appendix_fmnist_accuracy}
    % %\vspace{-4mm}
\end{figure}

As we can see in Figure~\ref{fig:diff_allocation_methods}, MIA and MDA fail to find the optimal decision, and thus their accuracy is inferior to ours. For MIA, even though the increasing number of participating clients in the later stage of model training will improve the model performance, the number of participating clients in the early stage of model training is extremely small, which leads to very inefficient model training and gradually deviates from the optimal model parameters. For MDA, since this method recruits a considerable number of participating clients at the early stage of model training, its model convergence rate is very fast. Even if the number of participating clients is small in the later stage, the model performance is not greatly affected. Yet, the accuracy of MIA and MDA is subject to experiment randomness. To better explain the deficiency of MIA and MDA, without loss of generality, we plot their test accuracy after each training round on the FMNIST dataset, which is shown in Figure~\ref{fig:appendix_fmnist_accuracy}. MIA is the worst one at the beginning stage (when $t<50$), while MDA cannot effectively improve model accuracy in the later stage (when $t>150$). That is why our algorithm outperforms MIA and MDA.

\section{Additional Experimental Results}
\label{appendix_case_study}

We implement FAIR \cite{deng2021fair} and FL-Market \cite{zheng2021incentive} which are designed for FL and DPFL,  respectively. For both FAIR and FL-Market, they investigated how to select clients based on their bids and data quality. They simply adopted EA for reward budget allocation by simply fixing the reward budget allocated to each communication round in advance. Other than implementing original FAIR and FL-Market, we also incorporate BARA into them to optimize the reward budget allocated to each communication round. For FL-Market, we randomly generate each client's privacy budget from a uniform distribution $U(0.1, 1)$ for MNIST, FMNIST and CIFAR-10 datasets. According to previous study, a privacy budget in $U(0.1, 1)$ can provide a very strong privacy protection. However,  for the CIFAR-100 dataset, the privacy budget of each client is sampled from a uniform distribution $U(10, 100)$ due to the fact that a high-dimensional model will be trained for classifying this complicated  dataset. Experimental results are presented in Table \ref{tab:case_study_fair} and Table \ref{tab:case_study_fl_market}. The results show that incorporating BARA into existing mechanisms can steadily improve the model accuracy by 1.33\% to 10.2\%  for all experiment scenarios. 
In other words, the BARA algorithm can generally improve model training performance by judiciously allocating rewards across multiple training rounds.  

\begin{table}[h]
% \large
\vspace{-2mm}
\setlength{\tabcolsep}{17.5pt} % Default value: 6pt
\renewcommand{\arraystretch}{1.3} % Default value: 1
\centering
\caption{The improvement of FAIR enhanced by BARA.}
\label{tab:case_study_fair}
\vspace{-2mm}
\begin{tabular}{l|c|c}
\hline
         & FAIR   & FAIR+BARA \\ \hline
MNIST    & 84.88 & \textbf{87.64}    \\ \hline
FMNIST   & 74.90 & \textbf{77.73}    \\ \hline
CIFAR-10  & 54.59 & \textbf{57.57}    \\ \hline
CIFAR-100 & 29.27 & \textbf{29.66}    \\ \hline
\end{tabular}
\vspace{-2mm}
\end{table}

\begin{table}[h]
% \large
% \vspace{-2mm}
\setlength{\tabcolsep}{11pt} % Default value: 6pt
\renewcommand{\arraystretch}{1.3} % Default value: 1
\centering
\caption{The improvement  of FL-Market enhanced BARA.}
\label{tab:case_study_fl_market}
\vspace{-2mm}
\begin{tabular}{l|c|c}
\hline
         & FL-Market   & FL-Market+BARA \\ \hline
MNIST    & 57.51 & \textbf{62.31}    \\ \hline
FMNIST   & 58.22 & \textbf{64.16}    \\ \hline
CIFAR-10  & 33.18 & \textbf{36.01}    \\ \hline
CIFAR-100 & 20.79 & \textbf{21.41}    \\ \hline
\end{tabular}
\vspace{-2mm}
\end{table}

\end{document}